\documentclass[lettersize,journal]{IEEEtran}
\usepackage{cite}
\usepackage{amsmath,amsfonts}
\usepackage{algorithmic}
\usepackage{array}
\usepackage[caption=false,font=normalsize,labelfont=sf,textfont=sf]{subfig}
\usepackage{textcomp}
\usepackage{stfloats}
\usepackage{url}
\usepackage{verbatim}
\usepackage{graphicx}
\usepackage{balance}
\usepackage{booktabs}
\usepackage{multirow}
\usepackage{xcolor}
\usepackage{cleveref}
\usepackage{bm}
\usepackage{amsmath}
\usepackage{siunitx}
\usepackage{algorithm}

\usepackage[switch, columnwise]{lineno}
\def\BibTeX{{\rm B\kern-.05em{\sc i\kern-.025em b}\kern-.08em
    T\kern-.1667em\lower.7ex\hbox{E}\kern-.125emX}}

\begin{document}

\title{Multi-Agent Dynamic Relational Reasoning for Social Robot Navigation}

\author{Jiachen Li$^{1*}$, Chuanbo Hua$^{2*}$, Jianpeng Yao$^{1}$, Hengbo Ma$^{3}$, Jinkyoo Park$^{2}$, \\ Victoria Dax$^{4}$, and Mykel J. Kochenderfer$^{4}$ 
    \thanks{* Equal contribution}
    \thanks{$^{1}$ J. Li and J. Yao are with the University of California, Riverside, CA, USA. {\tt\small \{jiachen.li,jyao073\}@ucr.edu}.}
    \thanks{$^{2}$ C. Hua and J. Park are with the Korea Advanced Institute of Science and Technology, Korea. {\tt\small \{cbhua,jinkyoo.park\}@kaist.ac.kr}.}
    \thanks{$^{3}$ H. Ma is with the University of California, Berkeley, CA, USA. {\tt\small hengbo\_ma@berkeley.edu}.}
    \thanks{$^{4}$ V. Dax and M. J. Kochenderfer are with Stanford University, CA, USA. {\tt\small \{vmdax,mykel\}@stanford.edu}.}
}

\maketitle

\begin{abstract}
Social robot navigation can be helpful in various contexts of daily life but requires safe human-robot interactions and efficient trajectory planning.
While modeling pairwise relations has been widely studied in multi-agent interacting systems, the ability to capture larger-scale group-wise activities is limited.
In this paper, we propose a systematic relational reasoning approach with explicit inference of the underlying dynamically evolving relational structures, and we demonstrate its effectiveness for multi-agent trajectory prediction and social robot navigation.
In addition to the edges between pairs of nodes (i.e., agents), we propose to infer hyperedges that adaptively connect multiple nodes to enable group-wise reasoning in an unsupervised manner.
Our approach infers dynamically evolving relation graphs and hypergraphs to capture the evolution of relations, which the trajectory predictor employs to generate future states.
Meanwhile, we propose to regularize the sharpness and sparsity of the learned relations and the smoothness of the relation evolution, which proves to enhance training stability and model performance.
The proposed approach is validated on synthetic crowd simulations and real-world benchmark datasets.
Experiments demonstrate that the approach infers reasonable relations and achieves state-of-the-art prediction performance. 
In addition, we present a deep reinforcement learning (DRL) framework for social robot navigation, which incorporates relational reasoning and trajectory prediction systematically. 
In a group-based crowd simulation, our method outperforms the strongest baseline by a significant margin in terms of safety, efficiency, and social compliance in dense, interactive scenarios.
We also demonstrate the practical applicability of our method with real-world robot experiments.
The code and experiment videos can be found on the project website: \url{https://relational-reasoning-nav.github.io/}.

\end{abstract}

\section{Introduction}

Social robot navigation has emerged as a crucial area of study in the rapidly evolving landscape of robotics, particularly in environments densely populated with humans (e.g., restaurants, hospitals, hotels). The integration of intelligent mobile robots into these shared spaces requires understanding human interactive behavior \cite{rudenko2020human,kothari2021human}. 
In the early literature, mobile robot navigation systems have been designed with a focus on static obstacle avoidance and efficiency, often neglecting the subtleties of human-robot interactions and social norms \cite{pandey2017mobile}. This oversight has limited their applicability in highly interactive human-centric environments.
As human-robot interactions and collaborations become increasingly common in recent years, it is inevitable to enable safe, efficient, and socially compliant robot navigation.

\begin{figure}[!tbp]
	\centering
	\includegraphics[width=0.9\columnwidth]{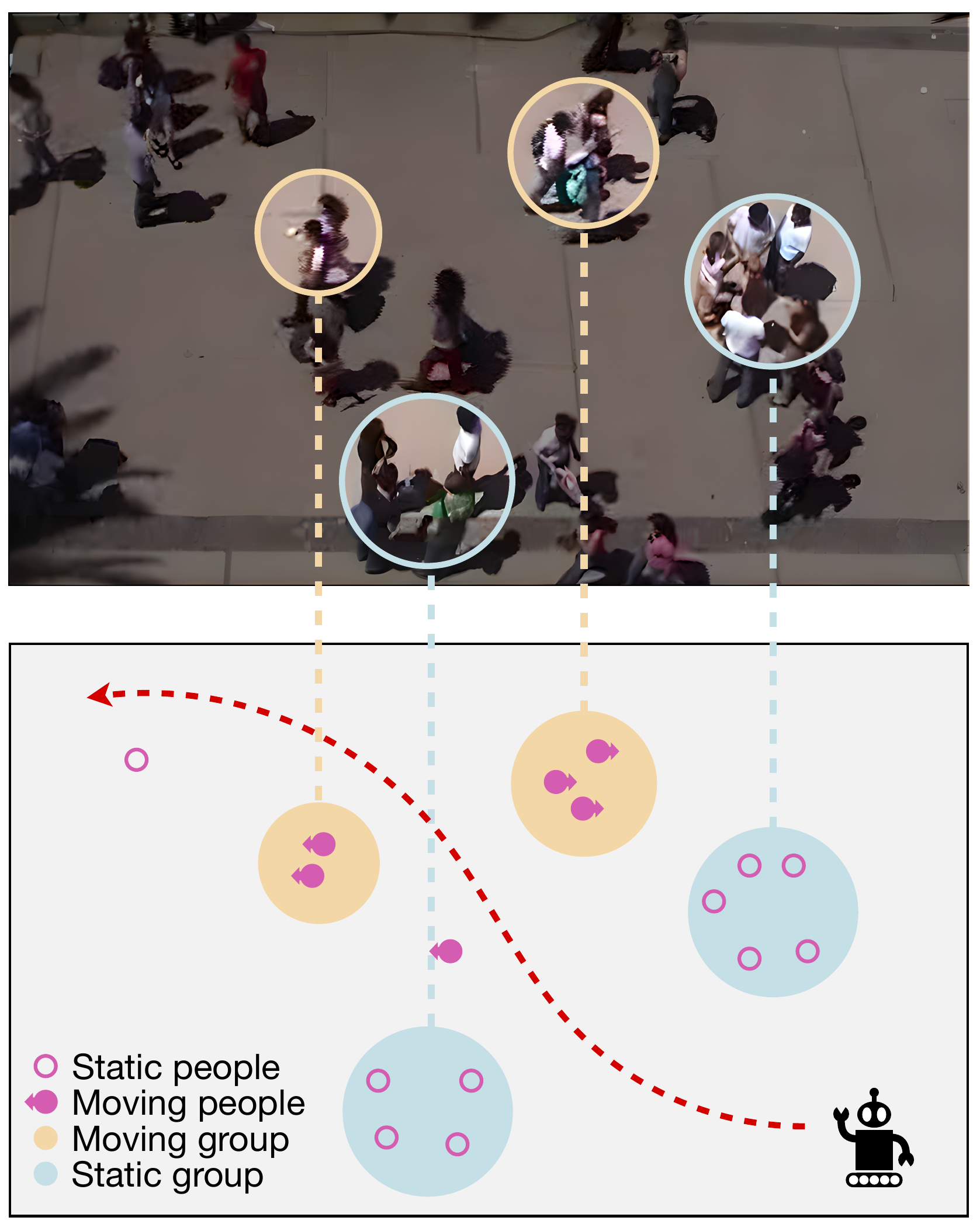}
	\caption{An illustrative example of social robot navigation in a dynamic, crowded scenario where the robot plans to reach its destination at the top left of the scene from its current location. Some pedestrians move individually while others behave similarly as a group. The relations between agents may evolve over time. The ellipses denote different groups of interacting agents that exhibit distinct group-wise relations, and the arrows indicate agents' motions. The robot needs to reach its destination safely and efficiently without colliding with pedestrians or intruding into group spaces.
    }
	\label{fig:teaser}
\end{figure}

A major challenge in social robot navigation lies in accurately modeling and predicting the future behavior of multi-agent interacting systems, which requires effective relational reasoning.
Fig. \ref{fig:teaser} provides a scenario that requires the robot to accurately understand both pairwise and group-wise relations between humans and infer how those relations influence their future behaviors.
Many research efforts in multi-agent trajectory prediction have been devoted to investigating pairwise relational reasoning \cite{kipf2018neural,li2020evolvegraph,graber2020dynamic,li2021rain} based on the graph representations where nodes and edges represent agents and their pairwise relations, respectively. 
However, group-wise relational reasoning and its implications in social navigation remain underexplored.

\begin{figure*}[!tbp]
	\centering
	\includegraphics[width=0.85\textwidth]{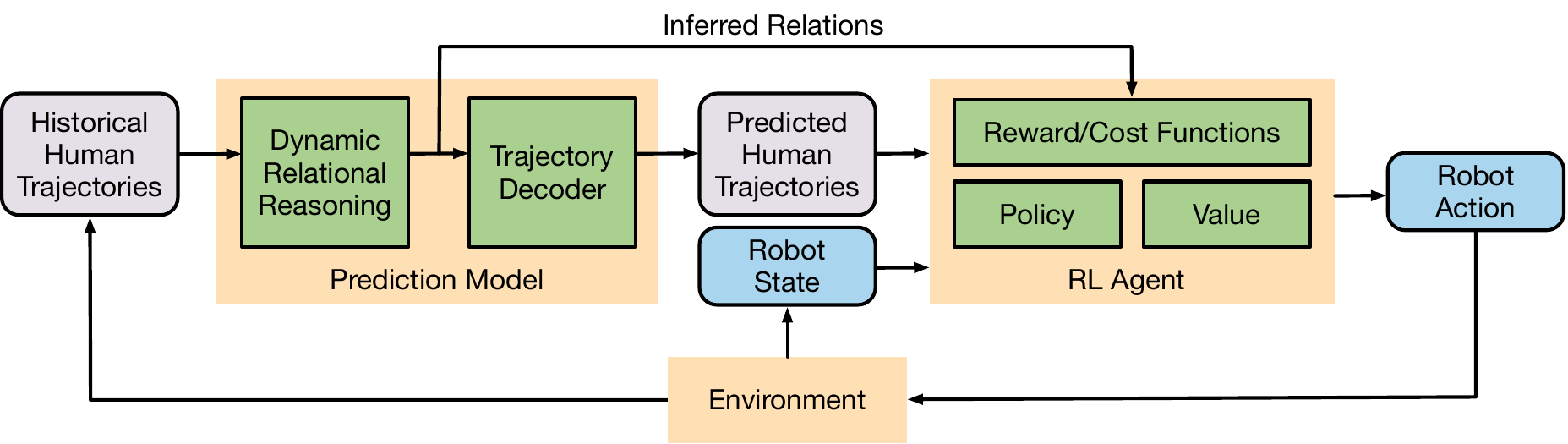}
	\caption{An overall diagram of our proposed pipeline, which consists of a trajectory prediction model and an RL agent for robot navigation. The prediction model takes in the historical observations of human motion and generates future trajectory hypotheses while inferring the underlying relations between humans. The RL agent decides the optimal robot actions based on the predicted trajectories and the current robot state. The inferred relations are used in designing the reward function. 
    We decouple the training of the prediction model and the RL agent into two sequential phases. 
    More details about the architectures and training strategy of the two components can be found in Sections \ref{sec:evolvehypergraph} and \ref{sec:social-robot-navigation}.}
	\label{fig:overall_diagram}
\end{figure*}

Although the recognition of group activities in images and videos has been extensively researched in the field of computer vision in recent years \cite{wu2021comprehensive}, these approaches primarily rely on the visual appearance information and human skeletal movements as informative cues for identifying group-level behaviors.
In contrast, only the historical state information is available in the context of trajectory prediction, which makes group recognition more difficult.
Some prior work proposed group identification methods that use heuristics based on the proximity of interactive agents \cite{bisagno2018group,zhou2022grouptron}, which work well in a specific scenario. However, it is difficult for these methods to directly generalize to diverse scenarios where the heuristics may need additional tuning.
For example, Group-LSTM \cite{bisagno2018group} uses the coherent filtering technique \cite{zhou2012coherent} to cluster the agents based on their collective movements. This group clustering strategy may work in scenarios where the agents in the same group behave similarly, but it may not generalize well to scenarios where agents in the same group can have highly distinct or even adversarial behaviors. Moreover, Group-LSTM models interactions implicitly by learning a feature embedding for each agent through a social pooling layer, which is fixed in the whole prediction horizon and difficult to explain.
Grouptron \cite{zhou2022grouptron} employs a spatio-temporal clustering technique based on the proximity of interactive agents and extracts interaction patterns at different scopes. However, group identification highly depends on the proximity threshold and cannot differentiate various types of relations.
GroupNet \cite{xu2022groupnet} is a more flexible approach for group inference and modeling based on graph neural networks. However, it cannot handle the dynamic evolution of relations over time and does not provide a way to leverage the inferred relations in social robot navigation.

To address these issues, we propose a systematic framework for trajectory prediction with dynamic pairwise and group-wise relational reasoning for multi-agent interacting systems, which extends our prior work \cite{li2020evolvegraph} that only handles pairwise relations.
A natural representation for group-wise relations is a hypergraph where each hyperedge can connect more than two nodes (i.e., agents).
Hypergraphs have been employed in areas such as chemistry, point cloud processing, and sensor network analysis.
Most existing works propose message passing mechanisms over hypergraphs with a pre-defined topology. 
In contrast, we propose an effective mechanism to infer the hypergraph topology for group-aware relational reasoning in a fully data-driven manner. Meanwhile, we propose a hypergraph evolution mechanism to enable dynamic reasoning about both pairwise and group-wise relations over time.
It is also necessary to encourage smooth evolutions of the relational structures and adjust their sharpness and sparsity in a flexible way for different applications. 

Moreover, we present a deep reinforcement learning (DRL) framework for social robot navigation adapted from CrowdNav++~\cite{liu2023intention} and a systematic way to employ our relational reasoning approach in the decision making process.
More specifically, we employ multi-head attention mechanisms to enable the robot to focus on important robot-human interactions and human-human interactions that may potentially influence the robot's planning, especially in highly crowded scenarios with a large number of agents.
Meanwhile, we employ the inferred relations between agents in the reward function design and incorporate the predicted future trajectories into the input of the policy network, which demonstrates the effectiveness of relational reasoning in social navigation.
The overall diagram of our proposed pipeline is illustrated in Fig. \ref{fig:overall_diagram}, which integrates the prediction model and reinforcement learning framework.

The main contributions of this paper are as follows:
\begin{itemize}
    \item We propose a systematic dynamic relational reasoning framework by inferring dynamically evolving latent interaction graphs and hypergraphs based on historical observations, and demonstrate its effectiveness for multi-agent trajectory prediction and social robot navigation. 
    \item We design a flexible mechanism to construct hyperedges and introduce a novel hypergraph attention mechanism to extract group features. We also propose new mechanisms to regularize the sharpness of learned relations, smoothness of relation evolution, and the sparsity of the learned graphs and hypergraphs, which not only enhance training stability but also reduce prediction error.
    \item We introduce a modified deep reinforcement learning framework for social robot navigation, which takes advantage of the inferred relations and predicted trajectories in a principled manner to improve safety and efficiency while complying with human social norms.
    \item We validate the trajectory prediction method with relational reasoning on both crowd simulations and real-world benchmark datasets. Our method infers reasonable evolving hypergraphs for group-wise relational reasoning and achieves state-of-the-art performance. We also evaluate the DRL-based robot navigation method on our group-based crowd simulations and real-world scenarios, demonstrating enhanced navigation performance, social awareness, and appropriateness in robot motion planning.
\end{itemize}

This paper builds upon our previous work \cite{li2020evolvegraph} in several important ways.
First, the relational reasoning in our earlier work only considers the fundamental pairwise relations while we additionally reason about group-wise relations in this work to enable systematic relational reasoning at different scales.
Second, we propose a series of effective regularization mechanisms to improve the learning of the dynamic evolution of relations in terms of smoothness, sharpness, and sparsity.
Third, our prior work only deals with the trajectory prediction problem while in this work we further investigate a systematic way to employ relational reasoning and trajectory prediction components in the downstream decision making task for social robot navigation. We also develop a group-based crowd simulator to evaluate our approach.

The remainder of the paper is organized as follows. Section~\ref{sec:related-work} reviews related work concisely. Section~\ref{sec:preliminaries} provides the necessary background related to the proposed approaches. Section~\ref {sec:evolvehypergraph} introduces the relational reasoning framework for trajectory prediction, and Section~\ref{sec:social-robot-navigation} introduces the deep reinforcement learning framework for social robot navigation. 
Section~\ref{sec:experiments} presents experimental settings, implementation details, and empirical results.
Finally, Section~\ref{sec:conclusion} concludes the paper by summarizing the key findings, discussing potential limitations, and suggesting directions for future research.

\section{Related work} \label{sec:related-work}

\subsection{Trajectory Prediction}
\label{sec: rw-trajpred}
The objective of trajectory prediction is to forecast a sequence of future states based on historical observations, which has been widely studied in dense scenarios with highly interactive entities (e.g., pedestrians \cite{rudenko2020human,kothari2021human,gu2022stochastic,bae2023set,bae2023eigentrajectory}, vehicles \cite{huang2022survey}, sports players \cite{kipf2018neural,li2020evolvegraph,graber2020dynamic,xu2022groupnet}).
Traditional approaches such as physics-based models \cite{helbing1995social,mehran2009abnormal}, heuristics-based models \cite{berg2011reciprocal,nilsson2015rule}, or probabilistic graphical models \cite{wang2011trajectory,schulz2018multiple} achieved satisfactory performance in simple scenarios where agents mostly move independently.
Recent advances in deep learning models have enhanced the capability of capturing not only the motion patterns of individuals but also the complex interactions between agents, which improves prediction accuracy. 
For instance, the temporal information of agent trajectories is captured by using sequential models such as long short-term memory (LSTM)\cite{alahi2016social, bisagno2018group}.
Multi-agent interaction behavior is modeled by graph neural networks (GNNs) \cite{salzmann2020trajectron++, zhou2022grouptron, xu2022groupnet}.
Also, generative models are incorporated in the prediction framework to capture the multi-modal distribution of future trajectories \cite{gupta2018social, kosaraju2019social,li2019conditional,salzmann2020trajectron++}.

Due to the promising performance of transformers in natural language processing \cite{devlin2018bert, radford2019language}, transformer-based approaches have also been prevailing in trajectory prediction \cite{yuan2021agentformer, giuliari2021transformer, ngiam2021scene, shi2022motion, MotionLM}.
For example, a unified transformer-based approach is proposed to consider different modalities such as road elements, agent interactions, and time steps \cite{ngiam2021scene}.
Shi et al. propose a transformer-based prediction framework that incorporates motion intention priors to improve prediction performance \cite{shi2022motion}.
An autoregressive transformer-based method called MotionLM generates joint predictions in a single autoregressive pass \cite{MotionLM}.
These approaches usually model multi-agent interactions implicitly while not providing explicit, explainable relational structures that can be used in downstream tasks.

\subsection{Social Robot Navigation}
Social robot navigation not only considers human behavior modeling but also robot motion planning \cite{mavrogiannis2023core}. The robot navigation system should consider the safety and efficiency of the robot and humans in the environment. 
Besides developing effective human behavior modeling methods for trajectory prediction to capture the interactions, another challenge is how to ensure the safety of humans during navigation. 
On one hand, some work focuses on game-theoretic approaches. 
For example, the human-robot interaction problem can be formulated as a non-cooperative non-zero-sum game \cite{turnwald2019human}. 
Multi-agent interaction problems can be formulated as a repeated linear-quadratic game \cite{fridovich2020efficient}.
On the other hand, some work proposes reinforcement learning based methods to implicitly capture the human-robot interactions and synthesize the robot planning policy directly based on the observations \cite{chen2017socially, chen2017decentralized, liu2020robot,katyal2022learning,liu2023intention}. 
Liu et al. \cite{liu2023intention} propose an RL-based social navigation method with trajectory prediction of surrounding pedestrians. However, their prediction and planning models do not consider group-wise relations or activities. 
Katyal et al. \cite{katyal2022learning} propose to learn a group-aware policy for robot navigation. However, their method only uses the historical state information for robot decision making while not considering the future behaviors of pedestrians. Also, their method requires ground truth group membership information during training and does not consider the evolution of group relations over time.
In contrast, our approach explicitly infers dynamic pairwise and group-wise relations in an unsupervised manner, which incorporates interaction modeling and human behavior modeling into the reinforcement learning framework systematically.

\subsection{Relational Learning and Reasoning}
The relational learning and reasoning module is crucial in multi-agent trajectory prediction \cite{rudenko2020human,huang2022survey}. It also shows effectiveness in social robot navigation \cite{mavrogiannis2023core}, physical dynamics modeling \cite{sanchez2020learning,battaglia2016interaction}, human-object interaction detection \cite{gao2020drg}, group activity recognition \cite{perez2022skeleton}, and community detection in various types of networks \cite{su2022comprehensive}.
The performance of these tasks highly depends on the learned relations or inferred interactions between different entities.
Traditional statistical relational learning approaches were proposed to infer the dependency between random variables based on statistics and logic, which are difficult to leverage high-dimensional data (e.g., images / videos) to infer complex relations \cite{koller2007introduction}.
Recently, deep learning has been investigated to model the relations or interactions such as pooling layers \cite{deo2018convolutional,alahi2016social}, attention mechanisms \cite{niu2021review,vemula2018social}, transformers \cite{yu2020spatio,giuliari2021transformer,yuan2021agentformer}, and graph neural networks \cite{battaglia2018relational,kipf2018neural,li2020evolvegraph,graber2020dynamic,li2021rain}, which can extract flexible relational features used in downstream tasks by aggregating the information of individual entities.
In contrast with most existing graph-based methods, which only use edges to model pairwise relations between agents, we propose to adopt effective hyperedges to additionally capture dynamic group-wise relations by inferring evolving hypergraphs.

Relational reasoning is an effective component for enabling safe and efficient robot navigation in dynamic environments. Model-free RL policies that use one-step predictions as observations can only capture instantaneous velocities, which are not sufficient for complex situations \cite{li2020socially, sathyamoorthy2020densecavoid}. 
While some studies predict the long-term goals of pedestrians to enhance planning, their reliance on a finite set of potential goals restricts generalizability \cite{katyal2020intent, vemula2017modeling}. Liu et al. \cite{liu2023intention} extend this by predicting pedestrian movement in a long-term, continuous two-dimensional space. However, their approach does not account for group-wise relations.
Our method considers both pairwise and group-wise relations to enhance the safety and social compliance of the learned robot policy.

\section{Preliminaries} \label{sec:preliminaries}

\subsection{Graph Neural Networks}
Graph Neural Networks (GNNs) are a class of deep learning models designed to capture the dependencies in graph-structured data.  Consider a graph $G = (V, E)$, where $V$ is the set of vertices (or nodes) and $E \subseteq V \times V$ is the set of edges. Each node $v \in V$ is associated with a feature vector $\mathbf{x}_v$. The goal of a GNN is to learn a state embedding $\mathbf{h}_v$ for each node that captures both its features and the neighbors' information given the structure of the graph.

The core idea behind GNNs is to update the representation of a node by aggregating the features of its neighbors:
\begin{equation}
    \mathbf{h}_v^{(k)} = g^{(k)}\left( \mathbf{x}_v, \text{AGG}^{(k)}\left( \{\mathbf{h}_u^{(k-1)} : u \in \mathcal{N}(v)\} \right) \right),
\end{equation}
where $\mathbf{h}_v^{(k)}$ is the feature vector of node $v$ at the $k$-th iteration, $\mathcal{N}(v)$ denotes the set of neighbors of $v$, and $\text{AGG}^{(k)}$ and $g^{(k)}$ are differentiable functions that aggregate and update node features, respectively.
In many GNN architectures, the aggregation function is a simple operation, such as sum, mean, or max. For instance, a common choice for the $\text{AGG}^{(k)}$ function is the mean aggregator:
\begin{equation}
    \text{AGG}^{(k)}(\cdot) = \frac{1}{|\mathcal{N}(v)|} \sum_{u \in \mathcal{N}(v)} \mathbf{h}^{(k-1)}_u,
\end{equation}
where $|\mathcal{N}(v)|$ is the number of nodes in the set of neighbors.
After $K$ iterations of aggregation, the final node representations $\mathbf{h}_v^{(K)}$ can be used for various graph-related tasks, such as node classification, link prediction, and graph classification. The effectiveness of GNNs lies in their ability to capture both the relational structure and node features in learning tasks.

\subsection{Markov Decision Process and Reinforcement Learning}
A Markov decision process (MDP) provides a mathematical framework used to model sequential decision making in discrete-time stochastic control \cite{bellman1957markovian}. 
An MDP is defined by a tuple \((S, A, P, R, \gamma)\) where:
\begin{itemize}
    \item \(S\) denotes a set of the agent's states.
    \item \(A\) denotes a set of actions available to the agent.
    \item \(P: S \times A \times S \rightarrow [0,1]\) denotes the transition function \(P(s_{t+1} \mid s_t, a_t)\), which defines the probability of transitioning from state \(s_t\) to \(s_{t+1}\) after taking action \(a_t\).
    \item \(R: S \times A \rightarrow \mathbb{R}\) denotes the reward function \(R(s_t, a_t)\).
    \item \(\gamma\) denotes the discount factor satisfying \(0 \leq \gamma \leq 1\).
\end{itemize}

The environment is defined by a set of states and actions, with transitions between states being probabilistic and dependent on the actions taken. The agent receives rewards or penalties after each action, guiding its learning process. 
A reinforcement learning agent learns to make decisions by interacting with an environment modeled as an MDP to maximize the expected discounted sum of rewards over time.

\subsection{Multi-Head Attention}
The attention mechanism \cite{vaswani2017attention} is an important feature of many state-of-the-art neural network architectures, particularly in handling sequential data such as language processing or time series analysis. It allows the model to focus on different parts of the input sequence simultaneously. 

The mechanism operates on queries (\(Q\)), keys (\(K\)), and values (\(V\)), which are typically derived from the same input in self-attention models. For a single attention head, we have
\begin{equation}
    \text{Attention}(Q, K, V) = \text{softmax}\left(\frac{QK^T}{\sqrt{d_k}}\right)V,
\end{equation}
where \(d_k\) is the dimension of keys, and the softmax function is applied row-wise. This formulation computes a set of attention weights and applies them to the values, effectively determining which parts of the input sequence are emphasized.
In the multi-head attention mechanism, this process is repeated across multiple heads, each with its own set of learned linear transformations for \(Q\), \(K\), and \(V\).

\section{Dynamic Relational Reasoning for \\Trajectory Prediction} \label{sec:evolvehypergraph}

Trajectory prediction involves forecasting the future trajectory of dynamic objects based on historical observations. We assume that, without loss of generality, there are $N$ agents in the scene. The number of agents may vary in different cases. 
We denote a set of agents' state sequences as 
\begin{equation}
    \mathbf{X}_{1:T} = \{\mathbf{x}^i_{1:T} \mid T=T_\text{h}+T_\text{f},i=1,\ldots,N\},
\end{equation}
where $N$ represents the number of involved agents, $T_\text{h}$ and $T_\text{f}$ represent the history and future prediction horizons, respectively. More specifically, we have $\mathbf{x}^i_t = (x^i_t,y^i_t)$ as the state of agent $i$, where $(x,y)$ represents the 2D coordinate.

The objective of probabilistic trajectory prediction is to estimate the conditional distribution for future trajectories given the historical observation $p(\mathbf{X}_{T_\text{h}+1:T_\text{h}+T_\text{f}} \mid \mathbf{X}_{1:T_\text{h}})$. As an intermediate step, we also aim to infer the underlying relational structures in the form of dynamic graphs (i.e., pairwise relations) and hypergraphs (i.e., group-wise relations), which may evolve over time and largely influence the predicted trajectories.  
More specifically, we aim to estimate the conditional distribution of the edge types and hyperedge types given their extracted features.
We treat ``no edge / hyperedge'' as a special type, which indicates no relation between the involved agents.
Long-term prediction can be obtained by iterative short-term predictions with a horizon of $T_\text{p}$ ($\leq T_\text{f}$).

\subsection{Overview}

\begin{figure*}[!tbp]
	\centering
	\includegraphics[width=\textwidth]{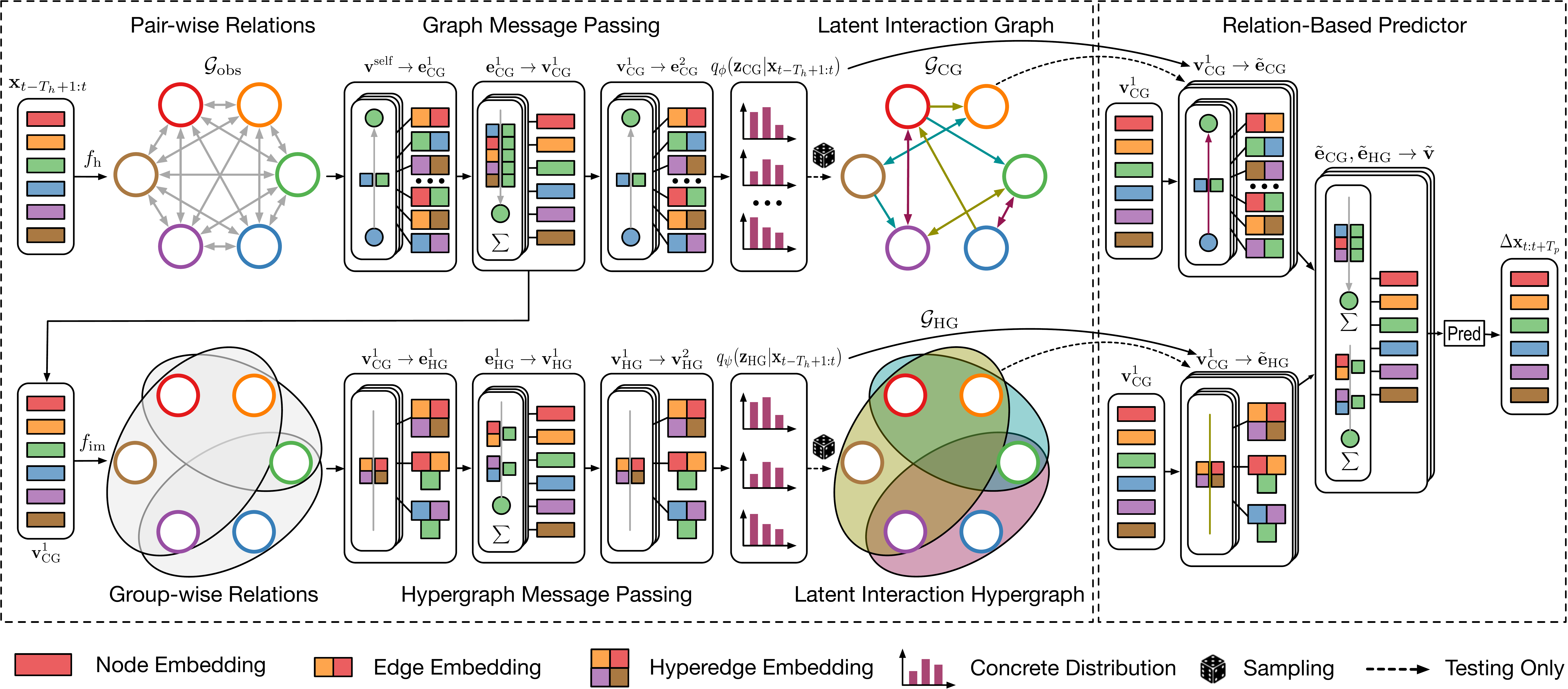}
	\caption{A diagram of our prediction method within a single prediction period, which consists of an encoder for relational reasoning and a decoder to generate prediction hypotheses. Different colors of edges in the latent interaction graph and hyperedges (i.e., ellipses) in the latent interaction hypergraph indicate different relation types. Best viewed in color.}
	\label{fig:model_diagram_static}
\end{figure*}

Fig. \ref{fig:model_diagram_static} shows the major components of our trajectory prediction approach, which follows an encoder-decoder architecture with an iterative sliding window to enable dynamic relational reasoning for long-term prediction. 
The encoder infers relation graphs and hypergraphs in the latent space based on historical state observations without direct supervision of relations. We introduce effective regularizations on the learned relations. The decoder employs the inferred latent relational structures to generate the distribution of future trajectory hypotheses. 
We present a dynamic reasoning mechanism to capture the probabilistic evolution of pairwise and group-wise relations.

We employ a fully connected observation graph $\mathcal{G}_\text{obs} = \{ \mathcal{V}_\text{obs}, \mathcal{E}_\text{obs}\}$ to represent the involved agents, where $\mathcal{V}_\text{obs} = \{ \mathbf{v}_i \mid i\in \{1,\ldots,N\} \}$ and $\mathcal{E}_\text{obs} = \{ \mathbf{e}_{ij} \mid i,j\in \{1,\ldots,N\} \}$. $\mathbf{v}_i$ and $\mathbf{e}_{ij}$ denote the attribute of node $i$ and the attribute of the directed edge from node $j$ to node $i$, respectively. Each agent node has two types of attributes: a \textit{self-attribute} encoding its own information and a \textit{social-attribute} encoding the aggregation of others' information.

\subsection{Encoder: Relational Reasoning}

The encoder consists of two parallel branches: \textit{pairwise} relational reasoning using a conventional graph ($\mathcal{G}_\text{CG}$) and \textit{group-wise} relational reasoning using a hypergraph ($\mathcal{G}_\text{HG}$), which are correlated via the node attributes obtained after a round of message passing across the observation graph $\mathcal{G}_\text{obs}$.

\textit{Computing pairwise edge features.} The self-attribute $\mathbf{v}^{\text{self}}_i$ and social attribute $\mathbf{v}^{\text{social}}_i$ of agent $i$ are obtained by
\begin{equation}\label{eq:graph_attention}
    \small
    \begin{split}
        \mathbf{v}^{\text{self}}_i &= \ f_\text{h} \left(\mathbf{x}^i_{t-T_h+1:t}\right), \\
	\alpha_{ij} &= \frac{\text{exp}(\text{LeakyReLU}(\mathbf{a}^{\top}[\mathbf{W}\mathbf{v}_i^{\text{self}}, \ \mathbf{W}\mathbf{v}_j^{\text{self}}]))}{\sum_{k\in\mathcal{N}_i}\text{exp}(\text{LeakyReLU}(\mathbf{a}^{\top}[\mathbf{W}\mathbf{v}_i^{\text{self}}, \ \mathbf{W}\mathbf{v}_k^{\text{self}}]))},\\
    \mathbf{e}^{1}_{\text{CG},ij} &= \ f^1_{\text{CG},e}\left(\left[\frac{\alpha_{ij}}{\alpha_{ij}+\alpha_{ji}}\mathbf{v}^{\text{self}}_i, \ \frac{\alpha_{ji}}{\alpha_{ij}+\alpha_{ji}}\mathbf{v}^{\text{self}}_j\right]\right), \\
    \mathbf{v}^{\text{social}}_i &= \ f^1_{\text{CG},v} \left( \sum_{j\in\mathcal{N}_i} \alpha_{ij}\mathbf{e}^{1}_{\text{CG},ij}\right), \ \sum_{j\in\mathcal{N}_i}\alpha_{ij} = 1,
    \end{split}
\end{equation}
where $f_\text{h}(\cdot)$ denotes the history embedding function, $\alpha_{ij}$ denotes the learnable attention weights, $\mathbf{e}^{1}_{\text{CG},ij}$, $f^1_{\text{CG},e}(\cdot)$ and $f^1_{\text{CG},v}(\cdot)$ denote the edge attribute, edge and node update functions during the first round of message passing across the conventional graph, respectively. Then we can obtain the complete node attribute of agent $i$ and conduct a second round of edge update by
\begin{equation}
    \small
    \begin{split}
        \mathbf{v}^1_{\text{CG},i} = \ \left[\mathbf{v}^{\text{self}}_i, \ \mathbf{v}^{\text{social}}_i\right],  \
	\mathbf{e}^{2}_{\text{CG},ij} = \ f^2_{\text{CG},e}\left(\left[\mathbf{v}^1_{\text{CG},i}, \ \mathbf{v}^1_{\text{CG},j}\right]\right).
    \end{split}
\end{equation}

\textit{Computing hyperedge features.} We infer the topology of the hypergraph $\mathcal{G}_\text{HG}$ in the form of incidence matrix $I_\text{HG}$ by 
\begin{equation}
    I_\text{HG} = f_\text{im} (\mathbf{V}^1_\text{CG}).
\end{equation}
Specifically, we first obtain a probabilistic incidence matrix by 
\begin{equation}
    I_\text{PIM} = f_\text{PIM} (\mathbf{V}^1_\text{CG}),
\end{equation}
where $I_{\text{PIM},im}$ is the probability of node $i$ being connected by hyperedge $\mathcal{H}_m$, the $i$-th row of the node attribute matrix $\mathbf{V}^1_\text{CG}$ is $\mathbf{v}^1_{\text{CG},i}$, and $f_{\text{PIM}}(\cdot)$ is a multi-layer perceptron (MLP). Then, we obtain the incidence matrix $I_{\text{HG}}$ by sampling the value of $I_{\text{HG},im}$ (1 or 0) from a Gumbel-Softmax distribution \cite{maddison2017concrete} defined by $I_{\text{PIM},im}$. 
In summary, we have 
\begin{equation}
    f_{\text{im}}(\mathbf{V}^1_\text{CG}):= \text{Gumbel-Softmax}(f_\text{PIM} (\mathbf{V}^1_\text{CG})).
\end{equation}
Note that we need to set the maximum number of hyperedges denoted as $M$ according to specific applications. Based on the inferred $I_\text{HG}$, the message passing across the hypergraph is given by
\begin{equation}\label{eq:hypergraph_attention}
    \small
    \begin{split}
        \alpha_{im} &= \frac{1}{N_{\mathcal{H}_m} - 1}\sum_{i\ne j, i, j\in\mathcal{H}_m}\alpha_{ij}, \\
    \mathbf{e}^{1}_{\text{HG},m} &= \ f^1_{\text{HG},e}\left(\sum_{i\in \mathcal{H}_m}\frac{\alpha_{im}}{\sum_{n:i\in\mathcal{H}_n}\alpha_{in}}\mathbf{v}^1_{\text{CG},i} \right),\\
    \alpha_{mi} &= \frac{\exp(\text{LeakyReLU}(\mathbf{a}^{\top}[\mathbf{W}_1\mathbf{e}_{\text{HG}, m}^1, \ \mathbf{W}_2\mathbf{v}_{\text{CG}, i}^1]))}{\sum_{j\in\mathcal{H}_m}\exp(\text{LeakyReLU}(\mathbf{a}^{\top}[\mathbf{W}_1\mathbf{e}_{\text{HG}, m}^1, \ \mathbf{W}_2\mathbf{v}_{\text{CG}, j}^1]))}, \\
    \mathbf{v}^1_{\text{HG},i} &= \ f^1_{\text{HG},v} \left( \sum_{m:i\in \mathcal{H}_m} \alpha_{mi}\mathbf{e}^{1}_{\text{HG},m} \right), \\ \mathbf{e}^{2}_{\text{HG},m} &= \ f^2_{\text{HG},e}\left(\sum_{i\in \mathcal{H}_m}\mathbf{v}^1_{\text{HG},i}\right),
    \end{split}
\end{equation}
where $\alpha_{im}$ denotes learnable attention weights from nodes to hyperedges, $N_{\mathcal{H}_m}$ denotes the number of nodes in the hyperedge $\mathcal{H}_m$, $\mathbf{e}^{1}_{\text{HG},m}$ denotes the attribute of hyperedge $\mathcal{H}_m$ ($m \in \{1,\ldots,M\}$), $\alpha_{mi}$ denotes learnable attention weights from hyperedges to nodes, $f^1_{\text{HG},e}(\cdot)$ / $f^2_{\text{HG},e}(\cdot)$ and $f^1_{\text{HG},v}(\cdot)$ denote the hyperedge and node update functions during the message passing across the hypergraph.

\textit{Inferring relation patterns.} The interaction graph $\mathcal{G}_\text{CG}$ and hypergraph $\mathcal{G}_\text{HG}$ represent relation patterns with a distribution of edge / hyperedge types, which are inferred based on the edge / hyperedge attributes. We denote the number of possible edge / hyperedge types as $L_\text{CG}$ and $L_\text{HG}$ which can be adjusted for different applications. Since not all pairs of agents have direct relations, we set the first type as ``no edge'' which prevents message passing through the corresponding edges and can be used to regularize the sparsity of relations. Formally, we formulate a classification problem which is to infer the edge / hyperedge (i.e., relation) types based on $\mathbf{e}^{2}_{\text{CG},ij}$ and $\mathbf{e}^{2}_{\text{HG},m}$ by
\begin{equation}
    \small
    \begin{split}
        \mathbf{z}_{\text{CG},ij} =& \ \text{Softmax}\left(\left(\mathbf{e}^2_{\text{CG},ij}+\mathbf{g}\right)/\tau\right), \ i,j\in \{1,\ldots,N\},\\
	\mathbf{z}_{\text{HG},m} =& \ \text{Softmax}\left(\left(\mathbf{e}^2_{\text{HG},m}+\mathbf{g}\right)/\tau\right), \ m\in \{1,\ldots,M\},
    \end{split}
\end{equation}
where $\mathbf{g}$ denotes a vector of i.i.d. samples from the Gumbel$(0,1)$ distribution and $\tau$ denotes the temperature coefficient which controls the smoothness of samples \cite{maddison2017concrete}. We can use the reparametrization trick to get gradients from this approximation, which is adapted from NRI \cite{kipf2018neural} and EvolveGraph \cite{li2020evolvegraph}. The latent variables $\mathbf{z}_{\text{CG},ij}$ and $\mathbf{z}_{\text{HG},m}$ are multinomial random vectors, which indicate the relation types for a certain edge / hyperedge. More formally, we have 
\begin{equation}
    \small
    \begin{split}
    \mathbf{z}_{\text{CG},ij} = \left[z_{\text{CG},ij,1},\ldots,z_{\text{CG},ij,L_\text{CG}}\right], \\
    \mathbf{z}_{\text{HG},m} = \left[z_{\text{HG},m,1},\ldots,z_{\text{HG},m, L_\text{HG}}\right],
    \end{split}
\end{equation}
where $L_\text{CG}$ and $L_\text{CG}$ denote the number of edge types and hyperedge types, respectively. Here, $z_{\text{CG},ij,r} \ (1\leq r \leq L_\text{CG})$ and $z_{\text{HG},m,s} \ (1\leq s \leq L_\text{HG})$ denote the probability of a certain edge / hyperedge belongs to a certain type, respectively. The operations in the encoder can be summarized as
\begin{equation}
    \small
    \begin{split}
        q_\text{CG}\left(\mathbf{z}_{\text{CG},\beta}\mid \mathbf{X}_{t-T_h+1:t}\right), \  q_\text{HG}\left(\mathbf{z}_{\text{HG},\beta}\mid \mathbf{X}_{t-T_h+1:t}\right) = \\ \text{Encoder}\left(\mathbf{X}_{t-T_h+1:t}\right).
    \end{split}
\end{equation}
\subsection{Decoder: Relation-Based Predictor}

\textit{Relational feature aggregation.} To enable dynamic relational reasoning, the whole prediction process consists of multiple recurrent prediction periods with a time horizon $T_\text{p}$ during which the relation graphs / hypergraphs are re-inferred. In a single prediction period, the decoder is designed to predict the distribution of future trajectories conditioned on the historical observations or the prediction hypotheses in the last prediction period as well as the relation graphs / hypergraphs inferred by the encoder, which is written as $p\left(\mathbf{X}_{t+1:t+T_\text{p}}\mid \mathbf{X}_{t-T_\text{h}+1:t}, \mathcal{G}_\text{CG}, \mathcal{G}_\text{HG}\right)$. We have
\begin{equation}
    \small
    \begin{split}
        &\Tilde{\mathbf{e}}_{\text{CG},ij} = \sum^{L_\text{CG}}_{l=1} z_{\text{CG},ij,l} \Tilde{f}^l_{\text{CG},e} \left(\left[\mathbf{v}^1_{\text{CG},i}, \mathbf{v}^1_{\text{CG},j}\right]\right),\\
	&\Tilde{\mathbf{e}}_{\text{HG},m} = \sum^{L_\text{HG}}_{l'=1} z_{\text{HG},m,l'} \Tilde{f}^{l'}_{\text{HG},e} \left( \sum_{i\in \mathcal{H}_m}\mathbf{v}^1_{\text{CG},i} \right), \label{eq:agg}\\
	&\Tilde{\mathbf{v}}_{i} = \Tilde{f}_v \left(\left[ \sum_{i\ne j}\Tilde{\mathbf{e}}_{\text{CG},ij}, 
	\sum_{i\in \mathcal{H}_m} \Tilde{\mathbf{e}}_{\text{HG},m}\right]\right),
    \end{split}
\end{equation}
where $\Tilde{\mathbf{e}}_{\text{CG},ij}$ and $\Tilde{\mathbf{e}}_{\text{HG},m}$ are, respectively, the aggregated edge / hyperedge attributes computed from the node attributes $\mathbf{v}^1_{\text{CG},i}$ obtained in the encoding process. 
The edge update functions corresponding to a certain edge / hyperedge type in $\mathcal{G}_\text{CG}$ and $\mathcal{G}_\text{HG}$ are denoted as $\Tilde{f}^l_{\text{CG},e}(\cdot)$ and $\Tilde{f}^{l'}_{\text{HG},e}(\cdot)$. 
The node update function in the decoding process is denoted as $\Tilde{f}_v(\cdot)$.

\textit{Predicting trajectory distributions.} To capture the uncertainty of future trajectories, the predicted distribution of displacement $\Delta\mathbf{x}^i_t$ at each time step is designed to be a Gaussian distribution with an equal constant covariance. The output function $\Tilde{f}_\text{output}(\cdot)$ takes in the updated node attribute $\Tilde{\mathbf{v}}_i$ and outputs a sequence of the mean of the Gaussian kernel of displacement $\Delta\mathbf{x}^i_{t:t+T_\text{p}-1}$ for agent $i$. We iteratively calculate the mean of the coordinates of agent $i$ in the future by
\begin{equation}
    \small
    \hat{\bm{\mu}}^{i}_{t+1} = \hat{\mathbf{x}}^i_t + \Delta\hat{\mathbf{x}}^i_t \ = \ \hat{\mathbf{x}}^i_t + \Tilde{f}_{\text{output},t}\left( \Tilde{\mathbf{v}}_i\right).
\end{equation}
Then, the predicted conditional distribution is obtained by
\begin{equation}
    \small
    p\left(\hat{\mathbf{x}}_{t+1}^{i}\mid \mathbf{z}_\text{CG}, \mathbf{z}_\text{HG}, \mathbf{x}_{t-T_h+1:t}^{i}\right) = \  \mathcal{N}\left(\hat{\mathbf{x}}_{t+1}^{i} \mid \hat{\bm{\mu}}^{i}_{t+1}, \sigma^2 \mathbf{I}\right),
\end{equation}
where $\sigma^2$ denotes the constant variance of Gaussian kernels, and $\mathbf{I}$ denotes the identity matrix.

\subsection{Dynamic Evolution of Relations}

We generalize the graph evolving mechanism proposed in our prior work \cite{li2020evolvegraph} to the dynamic inference of hypergraphs to capture evolving relations and the uncertainty of future trajectories. The operations include:
\begin{equation}
    \small
    \begin{split}
        q_\text{CG}\left(\mathbf{z}_{\text{CG},\beta}\mid \mathbf{X}_{1+\beta\tau:T_h+\beta\tau}\right)&,\ q_\text{HG}\left(\mathbf{z}_{\text{HG},\beta}\mid \mathbf{X}_{1+\beta\tau:T_h+\beta\tau}\right) \\
    &= \text{Encoder}\left(\mathbf{X}_{1+\beta\tau:T_h+\beta\tau}\right), 
    \end{split}
\end{equation}
\begin{equation}
    \small
    \begin{split}
        q_\text{CG}\left(\mathbf{z}'_{\text{CG},\beta}\mid \mathbf{X}_{1+\beta\tau:T_h+\beta\tau}\right) 
	= \text{GRU-CG}\left(q\left(\mathbf{z}_{\text{CG},\beta}\mid \mathbf{X}_{1+\beta\tau:T_h+\beta\tau}\right)\right),
    \end{split}
\end{equation}
\begin{equation}
    \small
    \begin{split}
        q_\text{HG}\left(\mathbf{z}'_{\text{HG},\beta}\mid \mathbf{X}_{1+\beta\tau:T_h+\beta\tau}\right) 
	=\text{GRU-HG}\left(q\left(\mathbf{z}_{\text{HG},\beta}\mid \mathbf{X}_{1+\beta\tau:T_h+\beta\tau}\right)\right),
    \end{split}
\end{equation}
where $\mathbf{z}'_{\text{CG},\beta}$ and $\mathbf{z}'_{\text{HG},\beta}$ denote the updated pairwise and group-wise relational structures, $\tau$ denotes the time gap between two consecutive relational inference steps, and $\beta$ denotes the index of relation graphs / hypergraphs starting from 0. Finally, the future trajectory distribution is
\begin{equation}
    \small
    \begin{split}
        p&\left(\mathbf{X}_{T_h+\beta\tau+1:T_h+(\beta+1)\tau}\mid \mathcal{G}'_{\text{CG},\beta},\mathcal{G}'_{\text{HG},\beta},\mathbf{X}_{1:T_h},\hat{\mathbf{X}}_{T_h+1:T_h+\beta\tau}\right)  \\  &= \ \text{Decoder}\left(\mathcal{G}'_{\text{CG},\beta},\mathcal{G}'_{\text{HG},\beta},\mathbf{X}_{1:T_h},\hat{\mathbf{X}}_{T_h+1:T_h+\beta\tau}\right).
    \end{split}
\end{equation}

\subsection{Regularizations on the Learned Relations}

We propose effective regularization techniques on learned underlying relational structures in three aspects: smoothness, sharpness, and sparsity. They influence the learned relations by shaping the edge and hyperedge type distributions.

First, both pairwise and group-wise relations tend to evolve gradually and smoothly over time. Thus, we propose a smoothness regularization loss that indicates the difference between two consecutive graphs or hypergraphs, which is written as
\begin{equation}
    \small
    \begin{split}
        L_\text{SM} =& \ \alpha_\text{SM, CG}\text{KL}\left( q_\text{CG}\left(\mathbf{z}'_{\text{CG},\beta}\right) \, || \, \ q_\text{CG}\left(\mathbf{z}'_{\text{CG},\beta+1}\right) \right) \\
	+& \ \alpha_\text{SM, HG}\text{KL}\left(q_\text{HG}\left(\mathbf{z}'_{\text{HG},\beta}\right) \,||\,  \ q_\text{HG}\left(\mathbf{z}'_{\text{HG},\beta+1}\right)\right),
    \end{split}
\end{equation}
where $\alpha_\text{SM, CG}$ and $\alpha_\text{SM, HG}$ denote the corresponding coefficients and KL$(\cdot)$ denotes the Kullback–Leibler divergence. The intuition is that the consecutive graphs / hypergraphs should have similar topologies in the form of edge / hyperedge type distributions.

Second, to learn different relation types that represent distinct relation patterns, the learned relation distributions should have an appropriate sharpness. To avoid the distributions being too close to uniform, we propose a sharpness regularization loss based on the entropy of edge / hyperedge type distributions, which is written as
\begin{equation}
    \small
    \begin{split}
        L_\text{SH} =
	-& \alpha_\text{SH, CG}q_\text{CG}\left(\mathbf{z}'_{\text{CG},\beta}\right) \log q_\text{CG}\left(\mathbf{z}'_{\text{CG},\beta}\right)  \\
	-& \alpha_\text{SH, HG} q_\text{HG}\left(\mathbf{z}'_{\text{HG},\beta}\right) \log q_\text{HG}\left(\mathbf{z}'_{\text{HG},\beta}\right),
    \end{split}
\end{equation}
where $\alpha_\text{SH, CG}$ and $\alpha_\text{SH, HG}$ denote the coefficients.

Third, human interactions tend to be sparse. They usually only interact with a subset of surrounding agents that may influence their behaviors, which implies that not all pairs of agents have a direct relation. Thus, we propose a sparsity regularization on the learned relations, which is written as
\begin{equation}
    \small
    \begin{split}
        L_\text{SP} &= \alpha_{\text{SP, CG}}\text{KL}\left(q_\text{CG}(\mathbf{z}'_{\text{CG}, \beta}) \ \| \ q_{\text{CG}, 0}\right)\\
        &+ \alpha_{\text{SP, HG}}\text{KL}\left(q_\text{HG}(\mathbf{z}'_{\text{HG}, \beta}) \ \| \ q_{\text{HG}, 0}\right)
    \end{split}
\end{equation}
where $\alpha_\text{SP, CG}$ and $\alpha_\text{SP, HG}$ denote the coefficients, $q_{\text{CG}, 0}$ and $q_{\text{HG}, 0}$ are categorical distributions where the probability of the ``no-relation'' type is $1$. This encourages sparser connections between nodes in the interaction graphs / hypergraphs.

\subsection{Training Strategy and Loss Function}

Since the inference of hyperedges highly depends on the intermediate node attributes $\mathbf{V}^1_\text{CG}$ obtained after the message passing across the observation graph $\mathcal{G}_\text{obs}$, we adopt a warm-up training stage during which only the modules related to the pairwise relational reasoning is trained. This training strategy provides a good initialization for the inference of hyperedges, which proves to accelerate convergence and improve the final performance. The complete loss at the formal training stage consists of a reconstruction loss, KL divergence losses, and regularization losses, written as
\begin{equation}
    \small
    \begin{split}
        &\qquad \qquad L = L_\text{Rec}  +  L_\text{KL} + L_\text{SM} + L_\text{SH}+L_\text{SP}, \\
	&\qquad \qquad \qquad L_\text{Rec} = \sum_{i=1}^{N}\sum_{t=T_\text{h}+1}^{T_\text{h}+T_\text{f}} \|\mathbf{x}^i_t - \bm{\mu}^i_t\|^2, \\
	&L_\text{KL} = \  \alpha_\text{KL, CG}\text{KL}\left( q_\text{CG}\left(\mathbf{z}'_{\text{CG},\beta}\mid\mathbf{X}_{1+\beta\tau:T_h+\beta\tau}\right) \,||\, \ p_\text{CG}\left(\mathbf{z}'_{\text{CG},\beta}\right)\right)   \\ 
	& \ \ \ \ \ + \ \alpha_\text{KL, HG}\text{KL}\left( q_\text{HG}\left(\mathbf{z}'_{\text{HG},\beta}\ | \ \mathbf{X}_{1+\beta\tau:T_h+\beta\tau}\right)\,||\, \ p_\text{HG}\left(\mathbf{z}'_{\text{HG},\beta}\right)\right),
    \end{split}
\end{equation}
where $p_\text{CG}\left(\mathbf{z}'_{\text{CG},\beta}\right)$ and $p_\text{HG}\left(\mathbf{z}'_{\text{HG},\beta}\right)$ denotes the prior uniform categorical distributions, and $\alpha_\text{KL, CG}$ and $\alpha_\text{KL, HG}$ denote the corresponding coefficients. The whole model can be trained in an end-to-end manner.
\section{Social Robot Navigation} \label{sec:social-robot-navigation}

\begin{figure*}[!tbp]
	\centering
	\includegraphics[width=\textwidth]{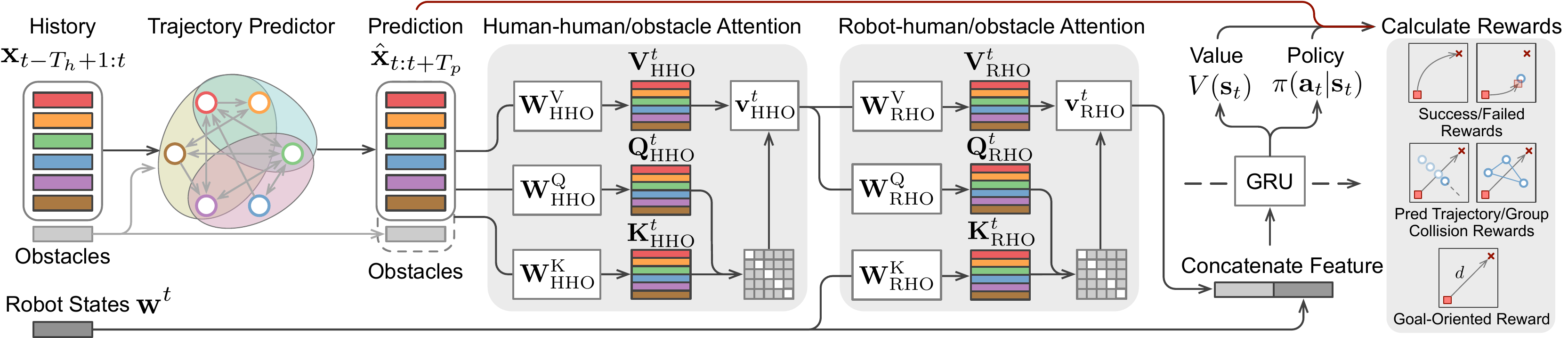}
    \vspace{-0.5cm}
	\caption{A diagram of the social robot navigation framework, which consists of a trajectory predictor, a human-human/obstacle attention layer, a robot-human/obstacle attention layer, a value network, and a policy network. Best viewed in color.}
    \vspace{-0.4cm}
	\label{fig:model_robot_navigation}
\end{figure*}

Social robot navigation entails determining a sequence of actions for a robot to reach its destination safely and efficiently based on the current robot's state, the behavior of surrounding pedestrians, and other contextual information. 
We formulate this problem as a Markov decision process \cite{liu2023intention}.
The state space \(S\) can include the robot's position, orientation, and the states of surrounding humans. 
The action space \(A\) may consist of different movement options. 
The transition probabilities \(P\) reflect the dynamics of the robot and its interactions with the environment, including the behavior of people around it. 
The reward function \(R\) is designed to encourage robot behaviors that are safe, efficient, and socially compliant.

In this section, we adopt the same notations for trajectories from Section \ref{sec:evolvehypergraph}. 
Given the historical positions of agent \(i\) (i.e., \(\mathbf{x}_{1:T_h}^i\)), we employ our method in Section \ref{sec:evolvehypergraph} to predict the future positions of the agent as \(\hat{\mathbf{x}}_{T_h+1:T_h+T_f}^i\). 
We denote the robot's states as $\mathbf{w}_t$, which includes current robot position \(\mathbf{x}_{R, t} = (x^{r}_t, y^{r}_t)\), the goal position \(\mathbf{g} = (g_x, g_y)\), the robot's velocity \(\mathbf{v}_t=(v_{x,t}, v_{y,t})\), the maximum robot speed $v_{\mathrm{max}}$, and the robot radius $\rho$. 
Furthermore, to enhance the realism and complexity of the environment, we also account for static obstacles as a global state, denoted as $\mathbf{O} = \{\mathbf{o}_k\}$. Each $\mathbf{o}_k$ represents the state of static obstacle $k$, which includes both its position and radius. In this work, we only consider circular static obstacles while more complex shapes are left for future work.
We define the state \(s_t \in \mathcal{S}\) of the MDP as 
\begin{equation}
    s_t = \left[ \mathbf{w}_t, \mathbf{x}_{T_\text{h}}, \hat{\mathbf{x}}_{T_\text{h}+1:T_\text{h}+T_\text{f}}, \mathbf{O}\right].
\end{equation}
The objective of social robot navigation is to learn an optimal robot policy $\pi(a_t \mid s_t)$, which navigates the robot to its destination efficiently within a certain time horizon without colliding with surrounding pedestrians or intruding into human personal spaces.

\subsection{Overview}

Fig. \ref{fig:model_robot_navigation} illustrates the overall pipeline of our robot navigation method, which integrates deep reinforcement learning with multi-agent relational reasoning and trajectory prediction. 
The pipeline consists of our trajectory prediction model, two attention modules (i.e., human-human/obstacle attention and robot-human/obstacle attention) for feature aggregation adapted from CrowdNav++ \cite{liu2023intention}, and a recurrent layer to output value and policy. 
Specifically, the trajectory predictor processes the historical trajectories of observable agents to generate their future state hypotheses. 
These predictions are first transformed by a human-human/obstacle attention module into a human-centric embedding. 
Then, these embeddings, along with the current robot state, are processed by the robot-human attention module, resulting in a refined feature representation of human agents considering human-robot interactions. This enhanced representation, concatenated with the current robot state, is then fed into a recurrent layer, which outputs the policy $\pi(a_t \mid {s}_t)$ and the value $V({s}_t)$.

\subsection{Attention Modules}
Attention layers in our model play an important role by assigning weights to each edge connecting to an agent, thereby emphasizing important interactions. The human-human/obstacle and robot-human/obstacle attention layers employ the scaled dot-product attention technique. 

In the human-human/obstacle attention layer, the current and predicted future states of humans are concatenated and passed through linear layers to derive the query, key, and value:
\begin{equation}
    (\mathbf{Q}_{\mathrm{HHO}}, \mathbf{K}_{\mathrm{HHO}}, \mathbf{V}_{\mathrm{HHO}}) = (W^\mathbf{Q}_{\mathrm{HHO}}\tilde{\mathbf{x}}, W^\mathbf{K}_{\mathrm{HHO}}\tilde{\mathbf{x}}, W^\mathbf{V}_{\mathrm{HHO}}\tilde{\mathbf{x}}),
\end{equation}
where \(\tilde{\mathbf{x}} = [\mathbf{x}_{T_h}, \hat{\mathbf{x}}_{T_h+1:T_h+T_f}]\) represents the concatenated state for all agents. The attention weights are obtained from \(\mathbf{Q}_{\mathrm{HHO}}\mathbf{K}_{\mathrm{HHO}}^\top\), and the human embeddings at the current time step $\mathbf{v}^t_{\mathrm{HHO}}$ are derived using a multi-head dot-product attention mechanism with eight attention heads.
Note that the attention weights between static obstacles are set to $0$ since they do not influence each other.

In the robot-human/obstacle attention layer, the human embeddings obtained from the human-human/obstacle attention layer serve as inputs for deriving the query and value:
\begin{equation}
    (\mathbf{Q}_{\mathrm{RHO}}, \mathbf{V}_{\mathrm{RHO}}) = (W^\mathbf{Q}_{\mathrm{RHO}}\mathbf{v}^t_{\mathrm{HHO}}, W^\mathbf{V}_{\mathrm{RHO}}\tilde{\mathbf{x}}\mathbf{v}^t_{\mathrm{HHO}}),
\end{equation}
while the key is computed using a linear embedding of the robot state:
\begin{equation}
    \mathbf{K}_{\mathrm{RHO}} = W^K_{\mathrm{RHO}}\mathbf{w}^t.
\end{equation}
The attention weights are calculated using \(\mathbf{Q}_{\mathrm{RHO}}\mathbf{K}_{\mathrm{RHO}}^T\), and the human embeddings at the current time after the second-layer attention \(\mathbf{v}^t_{\mathrm{RHO}}\) are obtained by applying a single-head dot-product attention mechanism.

\subsection{Policy and Value Estimation}
After obtaining the human embeddings, we concatenate them with the  robot embedding $h_{\mathrm{R}}^t = \mathrm{Linear}(\mathbf{w}^t)$, which are fed into a GRU layer \cite{cho2014learning}:
$${h}^t=\mathrm{GRU}(h^{t-1}, ([\mathbf{v}^t_{\mathrm{RHO}}, h_{\mathrm{R}}^t])),$$
where ${h}^t$ is the hidden state of the GRU at time $t$. 
The hidden state ${h}^t$ of the GRU is fed into a fully connected layer to obtain the value of the state $V({s}^t)$ and the policy $\pi(a_t \mid {s}_t)$.

\subsection{Reward Function}
The design of the reward function is crucial for social navigation based on reinforcement learning.
The primary objective of the robot is to reach a specified goal while adhering to social norms and avoiding collisions with individual pedestrians and groups. 
The components of the reward function are as follows:

\textit{Goal-oriented rewards.} To guide the robot toward its goal, we define a potential-based reward \( r_{\text{goal}} \), which is a function of the \( L2 \) distance between the robot's position \( \mathbf{x}_{R, t} \) and the goal position \( \mathbf{g} \) at time \( t \). Mathematically, it is written as
\begin{equation}
    r_{\text{goal}}(t) = -\alpha \cdot \|\mathbf{x}_{R, t} - \mathbf{g}\|_2,
\end{equation}
where \( \alpha \) is a scaling factor.
In addition, we introduce \( r_{\text{success}} \), a reward given when the robot successfully reaches the goal.

\textit{Human collision and prediction rewards.} To encourage safe navigation, we penalize collisions and potential intrusions into the space of humans. The prediction reward \( r_{\text{pred}\_\text{h}} \) penalizes the robot based on its proximity to the predicted future positions of humans, which is written as
\begin{equation}
    r_{\text{pred}
        \_\text{h}
    } = \beta
        _{1}
    \cdot \min_{\mathbf{p}_{\text{agent}} \in \mathbf{P}_{\text{pred}}}\left( \|\mathbf{x}_{R, t} - \mathbf{p}_{\text{agent}}\|_2 \right),
\end{equation}
where \( \beta
    _1
\) is a scaling factor, and \( \mathbf{P}_{\text{pred}} \) represents the set of predicted positions of agents.
The collision reward \( r_{\text{collision
    \_h
}} \) is given when the robot collides with any agents, which heavily penalizes such events.

\textit{Obstacle collision reward.} Similar to the human collision and prediction rewards, we have a distance reward \( r_{\text{dist\_o}} \) that penalizes the robot based on its proximity to the nearest obstacles, as expressed by:
\begin{equation}
    r_{\text{dist\_o}} = \beta_2 \cdot \min_{\mathbf{o}_k \in \mathbf{O}}\left( \|\mathbf{x}_{R, t} - \mathbf{o}_k\|_2 \right),
\end{equation}
where \( \beta_2 \) is a scaling factor. The collision reward \( r_{\text{coll\_o}} \) is applied when the robot collides with any obstacle, imposing a heavy penalty for such events. We set lower penalty weights for \( r_{\text{dist\_o}} \) and \( r_{\text{coll\_o}} \) compared to those associated with humans to emphasize safe navigation around humans.

\begin{figure}[!tbp]
	\centering
	\includegraphics[width=\columnwidth]{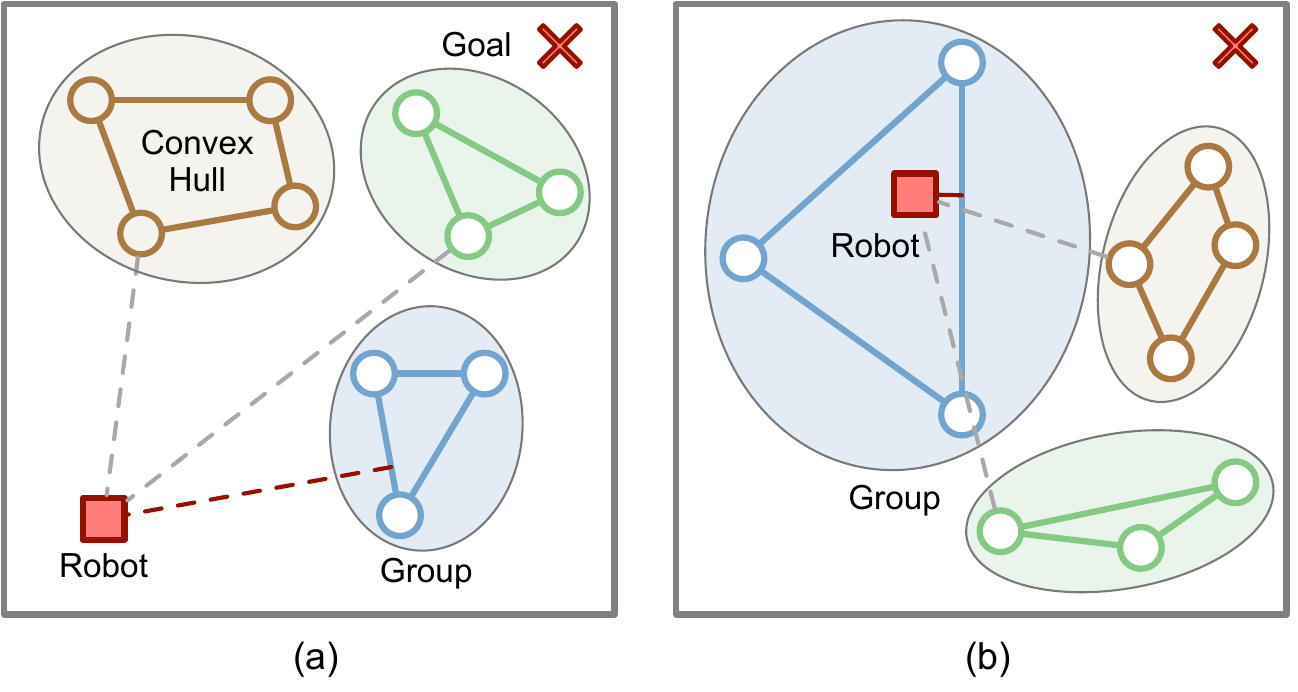}
    \vspace{-0.6cm}
	\caption{An illustrative visualization of two typical scenarios where the robot navigates in human crowds. (a) The robot is located outside the human groups' convex hulls, which complies with social norms. (b) The robot is located inside a human group's convex hull, which intrudes into human personal spaces and violates social norms. Best viewed in color.}
    \vspace{-0.2cm}
	\label{fig:group_reward}
\end{figure}

\textit{Group interaction reward.} To minimize discomfort to individuals and discourage intrusion into human groups, we consider the geometry of group formations. 
Fig. \ref{fig:group_reward} illustrates the scenarios where the robot complies with or violates the social norms.
For each group, we calculate a convex hull encompassing all agents within the group.
For example, with a group of pedestrian $G\lbrace\mathbf{x}^{i}\rbrace, i=1, \cdots, m$, where $m$ is the number of pedestrians in this group, we define the convex hull $H$ of this group as
$$H(G) = \bigcap \{C\ \mathrm{is\ a\ convex\ polygon}, G\subseteq C\},$$
i.e., the convex hull is the intersection of all convex sets that contain $G$. 
We employ the Qhull algorithm \cite{barber1996quickhull} to construct these convex hulls by initially selecting the extreme points based on $x$-coordinates and dividing the points into two subsets. Using a recursive divide-and-conquer approach, it identifies the farthest point from each line segment, refining the subsets and expanding the convex boundary until no external points remain. This method can efficiently delineate the smallest convex polygon enclosing all points.

The reward \( r_{\text{group}} \) is defined as a function of the robot's distance to the closest point on the group's convex hull, normalized by the area of the convex hull, which is written as
\begin{equation}
    r_{\text{group}} = \gamma \cdot \frac{c\ \|\mathbf{x}_{R, t} - H(\tilde{G})\|_2 }{\text{Area}(H(\tilde{G}))}, \tilde{G} = \mathrm{argmin}_{G}\|\mathbf{x}_{R, t} - H(\tilde{G})\|_2,
\end{equation}
here the $\|\mathbf{x}_{R, t} - H(\tilde{G})\|_2$ is the shortest distance from the robot to the convex hull, \( \gamma \) is a scaling factor. $c$ is a condition variable, which is set to $1$ when the robot is located outside the convex hull. Conversely, it is set to $-1$ to incentivize the robot to move out from the group with the shortest path.

\textit{Overall reward.} The overall reward function \( r({s}_t, {a}_t) \) at any time step \( t \) is defined as follows:
\begin{equation}\label{eq:reward}
    r(\mathbf{s}_t, \mathbf{a}_t) = 
        \begin{cases} 
        r_{\text{success}}, & \text{if }{s}_t\in S_{\text{goal}}, \\
        r_{\text{coll\_h}}, & \text{if }{s}_t\in S_{\text{fail\_h}}, \\
        r_{\text{coll\_o}}, & \text{if }{s}_t\in S_{\text{fail\_o}}, \\
        r_{\text{group}} + r_{\text{pred\_h}} + r_{\text{dist\_o}} + r_{\text{goal}}, & \text{otherwise},
        \end{cases}
\end{equation}
where $S_{\text{goal}}$ are the states of the robot reaching the goal, $S_{\text{fail\_h}}$ are the states of robot collision with a human, and $S_{\text{fail\_o}}$ are the states of robot collision with a static obstacle.

This reward design ensures that the robot receives a high reward when reaching the goal while also incorporating other critical aspects of social navigation, including collision avoidance and adherence to social norms during navigation.

\subsection{Training Strategy and Loss Function}

We train the trajectory prediction model and the reinforcement learning policy sequentially for the following reasons:
\begin{enumerate}
    \item \textit{Distinct Objectives:} The trajectory prediction model and the RL policy serve different purposes. The former is focused on accurately predicting human trajectories, which mainly considers pairwise and group-wise human-human interactions. On the other hand, the RL policy aims to optimize safe and efficient robot navigation behavior, which mainly considers robot-human interactions. An effective prediction model serves as a prerequisite for high-quality decision making for the robot. 
    \item \textit{Stability and Efficiency:} Joint training of both components can lead to instability and inefficiency due to the differing nature of the learning tasks, especially during the initial stage when the prediction model generates unreasonable future trajectories. Separating the training processes allows each component to learn more effectively from its specific objective function.
\end{enumerate}

To learn the RL policy, we employ Proximal Policy Optimization (PPO) \cite{schulman2017proximal}, a robust and efficient model-free policy gradient algorithm. 
\begin{algorithm}[!tbp]
\caption{Social Robot Navigation (Training Phase)}
\label{alg:social_robot_navigation}
\begin{algorithmic}[1]
\REQUIRE initial parameters of the reinforcement learning policy network $\theta_0$, clipping threshold $\epsilon = 0.2$
\FOR{$k = 1, 2, \dots, K$} 
    \STATE // Collect a trajectory of the robot by iteratively processing the action
    \FOR{$t = 1, 2, \dots, T$} 
        \STATE Predict future trajectories of pedestrians $\hat{\mathbf{x}}_{t+1:t+T_f}$ based on historical observations $\mathbf{x}_{t-T_{h}+1:t}$.
        \STATE Estimate the value of the state $V({s}_t)$ and the policy $\pi(a_t \mid {s}_t)$ with the current robot state, predicted pedestrians' trajectories, and static obstacles' states.
        \STATE Select an action using policy $\pi_k$ and obtain the next state and the reward from the environment.
    \ENDFOR
    \STATE // Update the learnable networks.
    \STATE Compute the rewards $r({s}_t, a_t)$.
    \STATE Estimate advantages $\hat{A}_t$.
    \STATE Iteratively update the policy by optimizing a surrogate objective function with PPO. Update the policy network by maximizing the PPO objective function via stochastic gradient ascent:
    \begin{align*}
        \theta_{k+1} = &\mathrm{argmax}_{\theta}\mathbb{E} \left[\min\left( \frac{\pi_{\theta}({a}_t\vert {s}_t)}{\pi_{\theta_k}({a}_t\vert {s}_t)}\hat{A}_t, \right.\right.\\
        &\left.\left. \text{clip}\left(\frac{\pi_{\theta}({a}_t\vert {s}_t)}{\pi_{\theta_k}({a}_t\vert {s}_t)}, 1 - \epsilon, 1 + \epsilon\right) \hat{A}_t \right) \right],
    \end{align*}
    where $ \theta $ represents the policy parameters, $ \hat{A}_t $ is an estimator of the advantage function at time $ t $, and $ \epsilon=0.2 $ is a hyperparameter denoting the clipping range.
\ENDFOR
\end{algorithmic}
\end{algorithm}
To enhance training stability and speed, we deploy 16 parallel environments, allowing the collection of diverse experiences. This parallelization facilitates efficient exploration of the state space. During policy updates, we use experiences collected over 30 steps from six episodes, providing a rich dataset for each update iteration.
The detailed training procedures are summarized in Algorithm \ref{alg:social_robot_navigation}.

\section{Experiments} \label{sec:experiments}

In this section, we validate the proposed approaches on benchmark datasets for trajectory prediction and a simulation for social robot navigation. The empirical results are analyzed and compared with state-of-the-art baselines.

\subsection{Benchmark Datasets and Crowd Simulation}

\subsubsection{Human Trajectory Datasets}
We evaluated human trajectory prediction and relational reasoning on three widely used real-world datasets and a synthetic human trajectory dataset. 

\textbf{ETH-UCY} \cite{pellegrini2009you,lerner2007crowds}: These datasets were collected in real-world traffic scenarios where pedestrians move on the street or the university campus. Some scenarios are crowded and full of pairwise and group-wise interactions between pedestrians who naturally form groups and move together. We used the standard split of training, validation, and testing datasets. Moreover, we also selected the test cases with more than eight pedestrians to construct a difficult test subset to compare the model performance in dense scenarios with strong interactions.

\textbf{SDD} \cite{robicquet2016learning}: This is a real-world pedestrian dataset captured on the university campus with top-down view videos with annotations on the positions of pedestrians. Most scenarios are crowded with interacting pedestrians and group activity. We used the standard split for training, validation, and testing.

\textbf{NBA}\footnote{\url{https://github.com/linouk23/NBA-Player-Movements}}: Although we focus on pedestrian trajectory prediction in this work, we also use a basketball player dataset to demonstrate the effectiveness of dynamic relational reasoning since the players have very complex interactions during the games. This dataset provides recordings of real NBA games that offer precise trajectory information of basketball players and the ball, which show complex team sports dynamics and feature high-intensity interactions among basketball players. The dataset is particularly suited for studying relational reasoning and group behavior dynamics in sports, highlighting the contrasting behaviors of cooperative team members and adversarial opponents.

These datasets enable a comprehensive evaluation of the model performance across various contexts.
The synthetic dataset is introduced below.

\subsubsection{Group-Aware Crowd Simulation}
We evaluate social robot navigation with a crowd simulation as an interactive environment for reinforcement learning, which is also used to collect a synthetic dataset. 
We develop a group-based crowd simulator that incorporates the ORCA model \cite{van2011reciprocal} to enable the simulation of crowd dynamics with human groups.
More specifically, we design a specific scenario to observe changes in group behaviors. 
The simulation environment is a 2D square space ($\SI{12}{m} \times \SI{12}{m}$), where we randomly initialize five groups of pedestrians. 
For both training and evaluation, each group contains 2-5 individuals randomly, i.e., in total 10-25 pedestrians. For each group, all members will be positioned within a circular area. Different groups do not have overlapping initial spaces.
Every group is assigned a random destination along the neighboring or the opposite boundary and moves at a random maximum speed (\(v_\text{max}\)) ranging from $\SI{0.5}{m/s}$ to $\SI{1.5}{m/s}$. Each pedestrian is modeled with a radius ($\rho$) between $\SI{0.3}{m}$ and $\SI{0.5}{m}$.
In the square, a random number of static obstacles, ranging from one to the number of groups, will be randomly positioned within a circular area. The center of this area coincides with the center of the square, and it has a radius of \SI{6}{m}, which is an inscribed circle. Each static obstacle is modeled with a radius between \SI{0.6}{m} and \SI{1.0}{m}.

For the robot simulation, during each episode, its starting and goal positions are randomly determined on the plane, ensuring a minimum distance of $\SI{10}{m}$ to avoid simple cases. The robot is equipped with a limited circular sensor range of $\SI{5}{m}$, reflecting a realistic hardware constraint. We assume the robot can move continuously at a maximum speed of $\SI{1}{m/s}$ and can instantly achieve its desired velocity.
To train and evaluate the trajectory prediction model, we generated 5k cases for training, 2k cases for validation, and 2k cases for testing.

\subsection{Evaluation Metrics}

\subsubsection{Trajectory Prediction}
To evaluate the trajectory prediction performance on both synthetic and benchmark datasets, we adopted two standard evaluation metrics \cite{li2020evolvegraph}:
\begin{itemize}
    \item \textbf{minADE}$_{\bm{20}}$: the minimum distance among $20$ trajectory samples between the predicted trajectories and the ground truth over all the entities in the prediction horizon.
    \item \textbf{minFDE}$_{\bm{20}}$: the minimum deviated distance among $20$ trajectory samples at the last predicted time step.
\end{itemize}

\subsubsection{Social Robot Navigation}
To assess the performance of social robot navigation, we adopt a comprehensive set of evaluation metrics \cite{mavrogiannis2023core}, which are designed to evaluate various aspects of social navigation, including safety, efficiency, and social appropriateness:
\begin{itemize}
    \item \textbf{Success Rate (SR)}: SR indicates the robot's capability to complete its navigation task within a certain period without collisions.
    \item \textbf{Collision Rate (CR)}: CR indicates navigation safety by quantifying how frequently it collides with humans.
    \item \textbf{Timeout Rate (TR)}: TR tracks the proportion of navigation tasks where the robot fails to reach its destination within a certain period without collisions, which indicates navigation efficiency.
    \item \textbf{Navigation Time (NT)}: NT shows the average duration, in seconds, for the robot to complete a navigation task successfully, which indicates navigation efficiency.
    \item \textbf{Path Length (PL)}: PL shows the length of the robot's path from its initial position to the goal, which indicates the navigation efficiency.
    \item \textbf{Intrusion Time Ratio (ITR)}: ITR is defined as $c/C$ where $c$ is the number of timesteps of the robot intrudes into any human's future position from $t+1$ to $t+5$, and $C$ is the length of that episode. The ITR assesses the robot's respect for socially sensitive spaces.
    \item \textbf{Social Distance (SD)}: SD is defined as the average distance between the robot and its closest human when an instruction to human spaces occurs, which gauges the robot's ability to maintain appropriate distances from individuals or groups, reflecting its social awareness.
    \item \textbf{Group Intersections (GI)}: GI shows the ratio of scenarios where the robot intrudes into human groups, indicating the social appropriateness of robot navigation.
\end{itemize}

\subsection{Baselines}

\subsubsection{Trajectory Prediction}
We compare our approach with a series of state-of-the-art baselines:
\begin{itemize}
    \item \textbf{MID} \cite{gu2022stochastic}: This is a stochastic trajectory prediction framework with motion indeterminacy diffusion, which contains a transformer-based architecture to capture the temporal dependencies in trajectories.
    \item \textbf{GroupNet} \cite{xu2022groupnet}: This trajectory prediction model learns to capture multi-scale interaction patterns based on graph representations and infers interaction strengths.
    \item \textbf{Graph-TERN} \cite{bae2023set}: This is a graph-based pedestrian trajectory estimation and refinement network, which consists of a control point prediction module, a trajectory refinement module, and a multi-relational graph convolution.
    \item \textbf{EigenTrajectory} \cite{bae2023eigentrajectory}: This is a state-of-the-art trajectory prediction approach that uses a trajectory descriptor to form a compact space in place of Euclidean space for representing pedestrian movements.
\end{itemize}

\subsubsection{Social Robot Navigation} 
To demonstrate the effectiveness of dynamic relational reasoning in social robot navigation, we compare our method with both traditional and state-of-the-art baselines:
\begin{itemize}
    \item \textbf{ORCA} \cite{van2011reciprocal}: This is a rule-based approach to reciprocal collision avoidance, where multiple independent agents avoid collisions with each other without communication among agents while moving in a common workspace.
    \item \textbf{DSOA} \cite{huber2022avoiding}: This is a closed-form solution for dynamic spatial navigation, which creates smooth motion fields to ensure the robot navigates safely within confined spaces and around both static and dynamic obstacles.
    \item \textbf{RL (no prediction)}: This is a standard reinforcement learning method without predicting the future behavior of surrounding humans. The observations only include past and current information. 
    \item \textbf{RL (constant velocity)}: This is a reinforcement learning method with constant velocity prediction of human trajectories. While considering the future information, this method does not model the complex interactions.
    \item \textbf{Group-Aware} \cite{katyal2022learning}: This model employs deep reinforcement learning to develop navigation policies that recognize and respect dynamic human group formations, which enhances robot navigation by reducing collisions and adhering to social norms.
    \item \textbf{CrowdNav++} \cite{liu2023intention}: This is a state-of-the-art social robot navigation method based on reinforcement learning. It infers the intentions of moving pedestrians, which prevents the robot from intruding into the humans' intended paths.
\end{itemize}

\begin{table*}[!tbp]
	\setlength{\tabcolsep}{1mm}
	\caption{minADE$_{20}$ / minFDE$_{20}$ (meters) on the ETH / UCY datasets (overall).}
	\vspace{-0.2cm}
	\fontsize{8}{8}\selectfont
	\resizebox{\textwidth}{!}{
		\begin{tabular}{m{0.9cm}<{\centering}| m{1.3cm}<{\centering} m{1.3cm}<{\centering} m{1.3cm}<{\centering} m{1.5cm}<{\centering} | m{1.3cm}<{\centering} m{1.3cm}<{\centering} m{1.3cm}<{\centering} m{1.3cm}<{\centering} m{1.3cm}<{\centering} m{1.3cm}<{\centering} m{1.3cm}<{\centering} m{1.3cm}<{\centering}}
			\toprule
			\midrule
			\multirow{3}*{\shortstack[lb]{}} 
			& \multicolumn{4}{c|}{Baseline Methods} & \multicolumn{8}{c}{\textbf{Ours}} \\
			\cline{2-13} 
			\vspace{0.8cm}
			Dataset   &  MID \cite{gu2022stochastic} & GroupNet \cite{xu2022groupnet}  & Graph-TERN \cite{bae2023set}  & EigenTraj \cite{bae2023eigentrajectory} & SCG & SHG & SCG + SHG & DCG + DHG & DCG + DHG + SM & DCG + DHG + SM + SH & DCG + DHG + SM + SH + SP (Equal Attention) & DCG + DHG + SM + SH + SP \\
			\midrule 
			ETH & 0.39 / 0.66 & 0.46 / 0.73 & 0.42 / 0.58 & 0.36 / 0.57 & 0.41 / 0.68 & 0.45 / 0.76 & 0.40 / 0.66 & 0.37 / 0.63 & 0.35 / 0.62 & \textbf{0.33} / 0.59 & 0.34 / 0.60 & \textbf{0.33} / \textbf{0.54} \\
			HOTEL & 0.13 / 0.22 & 0.15 / 0.25 & 0.14 / 0.23 & 0.13 / 0.21 & 0.16 / 0.27 & 0.19 / 0.28 & 0.16 / 0.24 & 0.15 / 0.23 & 0.13 / 0.21 & 0.13 / 0.18 & 0.14 / 0.21 & \textbf{0.12} / \textbf{0.17} \\
			UNIV & 0.22 / 0.45 & 0.26 / 0.49 & 0.26 / 0.45 & 0.24 / 0.43 & 0.24 / 0.49 & 0.28 / 0.51 & 0.22 / 0.47 & 0.21 / 0.45 & 0.20 / 0.43 & \textbf{0.18} / 0.41 & 0.19 / 0.40 & \textbf{0.18} / \textbf{0.36} \\
			ZARA1 & \textbf{0.17} / 0.30 & 0.21 / 0.39 & 0.21 / 0.37 & 0.20 / 0.35 & 0.23 / 0.36 & 0.25 / 0.40 & 0.22 / 0.35 & 0.21 / 0.32 & 0.20 / 0.31 & 0.18 / 0.30 & 0.19 / 0.29 & \textbf{0.17} / \textbf{0.29} \\
			ZARA2 & \textbf{0.13} / 0.27 & 0.17 / 0.33 & 0.17 / 0.29 & 0.14 / 0.25 & 0.21 / 0.30 & 0.23 / 0.31 & 0.20 / 0.27 & 0.18 / 0.24 & 0.17 / 0.20 & 0.15 / 0.21 & 0.16 / 0.23 & \textbf{0.13} / \textbf{0.20} \\
			\midrule
			AVG & 0.21 / 0.38 & 0.25 / 0.44 & 0.24 / 0.38 & 0.21 / 0.36 & 0.24 / 0.45 & 0.26 / 0.47 & 0.23 / 0.42 & 0.21 / 0.39 & 0.20 / 0.35 & 0.19 / 0.32 & 0.21 / 0.36 & \textbf{0.18} / \textbf{0.30} \\
			\bottomrule
		\end{tabular}
	}
	\label{tab:ETH_overall}
\end{table*}

\begin{table*}[!tbp]
	\setlength{\tabcolsep}{1mm}
	\caption{minADE$_{20}$ / minFDE$_{20}$ (meters) on the ETH / UCY datasets (difficult subset).}
	\vspace{-0.2cm}
	\fontsize{8}{8}\selectfont
	\resizebox{\textwidth}{!}{
		\begin{tabular}{m{0.9cm}<{\centering}| m{1.3cm}<{\centering} m{1.3cm}<{\centering} m{1.3cm}<{\centering} m{1.5cm}<{\centering} | m{1.3cm}<{\centering} m{1.3cm}<{\centering} m{1.3cm}<{\centering} m{1.3cm}<{\centering} m{1.3cm}<{\centering} m{1.3cm}<{\centering} m{1.3cm}<{\centering} m{1.3cm}<{\centering}}
			\toprule
			\midrule
			\multirow{3}*{\shortstack[lb]{}} 
			& \multicolumn{4}{c|}{Baseline Methods} & \multicolumn{8}{c}{\textbf{Ours}} \\
			\cline{2-13} 
			\vspace{0.8cm}
			Dataset   &  MID \cite{gu2022stochastic} & GroupNet \cite{xu2022groupnet}  & Graph-TERN \cite{bae2023set}  & EigenTraj \cite{bae2023eigentrajectory} & SCG & SHG & SCG + SHG & DCG + DHG & DCG + DHG + SM & DCG + DHG + SM + SH & DCG + DHG + SM + SH + SP (Equal Attention) & DCG + DHG + SM + SH + SP \\
			\midrule 
			ETH & 0.42 / 0.77 & 0.49 / 0.89 & 0.52 / 0.84 & 0.40 / 0.73 & 0.45 / 0.79 & 0.54 / 0.96 & 0.43 / 0.72 & 0.41 / 0.71 & 0.38 / 0.65 & 0.36 / 0.63 & 0.36 / 0.65 & \textbf{0.35} / \textbf{0.61} \\
			HOTEL & 0.15 / 0.25 & 0.18 / 0.30 & 0.17 / 0.28 & \textbf{0.14} / 0.24 & 0.20 / 0.31 & 0.24 / 0.35 & 0.17 / 0.29 & 0.17 / 0.26 & 0.15 / 0.24 & \textbf{0.14} / 0.23 & 0.15 / 0.25 & \textbf{0.14} / \textbf{0.20} \\
			UNIV & 0.23 / 0.55 & 0.27 / 0.58 & 0.29 / 0.61 & 0.26 / 0.54 & 0.28 / 0.55 & 0.33 / 0.69 & 0.26 / 0.54 & 0.24 / 0.51 & 0.23 / 0.48 & 0.22 / 0.46 & 0.22 / 0.49 & \textbf{0.21} / \textbf{0.45} \\
			ZARA1 & 0.21 / 0.38 & 0.26 / 0.45 & 0.23 / 0.42 & 0.22 / 0.37 & 0.25 / 0.41 & 0.29 / 0.52 & 0.23 / 0.38 & 0.23 / 0.36 & 0.21 / 0.35 & 0.20 / 0.33 & 0.21 / 0.33 & \textbf{0.19} / \textbf{0.31} \\
			ZARA2 & 0.20 / 0.34 & 0.21 / 0.36 & 0.21 / 0.32 & 0.18 / 0.28 & 0.26 / 0.32 & 0.28 / 0.41 & 0.23 / 0.29 & 0.21 / 0.28 & 0.19 / 0.25 & 0.18 / 0.24 & 0.19 / 0.26 & \textbf{0.16} / \textbf{0.23} \\
			\midrule
			AVG & 0.31 / 0.50 & 0.29 / 0.49 & 0.27 / 0.47 & 0.25 / 0.44 & 0.30 / 0.56 & 0.34 / 0.65 & 0.28 / 0.50 & 0.25 / 0.45 & 0.23 / 0.41 & 0.22 / 0.39 & 0.24 / 0.43 & \textbf{0.20} / \textbf{0.34} \\
			\bottomrule
		\end{tabular}
	}
	\label{tab:ETH_difficult}
\end{table*}

\begin{table*}[!tbp]
	\setlength{\tabcolsep}{1mm}
	\caption{minADE$_{20}$ / minFDE$_{20}$ on the SDD (pixels) and NBA (meters) datasets.}
	\vspace{-0.2cm}
	\fontsize{8}{8}\selectfont
	\resizebox{\textwidth}{!}{
		\begin{tabular}{m{0.9cm}<{\centering}| m{1.3cm}<{\centering} m{1.3cm}<{\centering} m{1.3cm}<{\centering} m{1.5cm}<{\centering} | m{1.3cm}<{\centering} m{1.3cm}<{\centering} m{1.3cm}<{\centering} m{1.3cm}<{\centering} m{1.3cm}<{\centering} m{1.3cm}<{\centering} m{1.3cm}<{\centering} m{1.3cm}<{\centering}}
			\toprule
			\midrule
			\multirow{3}*{\shortstack[lb]{}} 
			& \multicolumn{4}{c|}{Baseline Methods} & \multicolumn{8}{c}{\textbf{Ours}} \\
			\cline{2-13} 
			\vspace{0.8cm}
			Dataset   &  MID \cite{gu2022stochastic} & GroupNet \cite{xu2022groupnet}  & Graph-TERN \cite{bae2023set}  & EigenTraj \cite{bae2023eigentrajectory} & SCG & SHG & SCG + SHG & DCG + DHG & DCG + DHG + SM & DCG + DHG + SM + SH & DCG + DHG + SM + SH + SP (Equal Attention) & DCG + DHG + SM + SH + SP \\
			\midrule 
			SDD  & 9.7 / 15.3 & 9.7 / 15.3 & 8.4 / 14.3 & \textbf{8.1} / 13.1 & 12.7 / 22.5 & 14.0 / 25.8 & 12.0 / 20.4 & 9.5 / 15.4 & 8.8 / 14.7 & 8.4 / 13.6 & 9.4 / 14.9 & \textbf{8.1} / \textbf{12.8} \\
			NBA & 1.65 / 2.66 & 1.53 / 2.45  & 1.46 / 2.19 & 1.49 / 2.30 & 1.69 / 2.73 & 1.80 / 3.04 & 1.56 / 2.32 & 1.37 / 2.00 & 1.28 / 1.81 & 1.23 / 1.73 & 1.31 / 1.87 & \textbf{1.16} / \textbf{1.64} \\
			\bottomrule
		\end{tabular}
	}
	\vspace{-0.3cm}
	\label{tab:SDD_NBA}
\end{table*}

\subsection{Implementation Details}

\subsubsection{Trajectory Prediction}
We use a batch size of 32 and an initial learning rate of 0.001. The model is trained for 10k epochs with an Adam optimizer and a step scheduler with a decay rate of 0.85. We set the weight of SM, SH, and SP to 0.001. 
The details of the model components are below: 
\begin{itemize}
    \item \textit{Agent node embedding function}: a three-layer MLP with hidden size = 128.
    \item \textit{Inferring hypergraph topology function}: a three-layer MLP with hidden size = 128.
    \item \textit{Edge feature updating function}: a distinct three-layer MLP with hidden size = 128 for each type of edge or hyperedge.
    \item \textit{Encoding}: a three-layer MLP with hidden size = 128.
    \item \textit{Decoding}: a three-layer MLP with hidden size = 128.
    \item \textit{Recurrent graph / hypergraph evolution module}: a two-layer GRU with hidden size = 128.
\end{itemize}

Specific experimental details of different datasets:
\begin{itemize}
    \item \textit{Synthetic}: 3 edge / hyperedge types, maximum hyperedge = 5, prediction period = 1, encoding horizon = 4. 
    \item \textit{ETH-UCY}: 3 edge / hyperedge types, maximum hyperedge = 5, prediction period = 4, encoding horizon = 8.
    \item \textit{SDD}: 5 edge / hyperedge types, maximum hyperedge = 8, prediction period = 4, encoding horizon = 8.
    \item \textit{NBA}: 3 edge / hyperedge types, maximum hyperedge = 5, prediction period = 5, encoding horizon = 5.
\end{itemize}

\subsubsection{Social Robot Navigation} 

To simulate the behaviors of the robot and pedestrians, we employ holonomic kinematics. The action space for each agent, including the robot, at time $t$ is represented by their desired velocities along the $x$ and $y$ axes, denoted as $a_t = [v_{x,t}, v_{y,t}]$. 
This action space is continuous, with the robot's maximum speed at $\SI{1}{m/s}$. 
Humans are controlled by ORCA \cite{van2011reciprocal} and interact with surrounding humans and obstacles. 
We operate under the assumption that all agents can instantaneously achieve their desired velocities and maintain these velocities for a duration of $\Delta t$ seconds. The position of an agent at any time is updated based on the velocities according to the update rule, which ensures realistic motion patterns for the robot and human agents.

\begin{figure*}[!tbp]
	\centering
	\includegraphics[width=\textwidth]{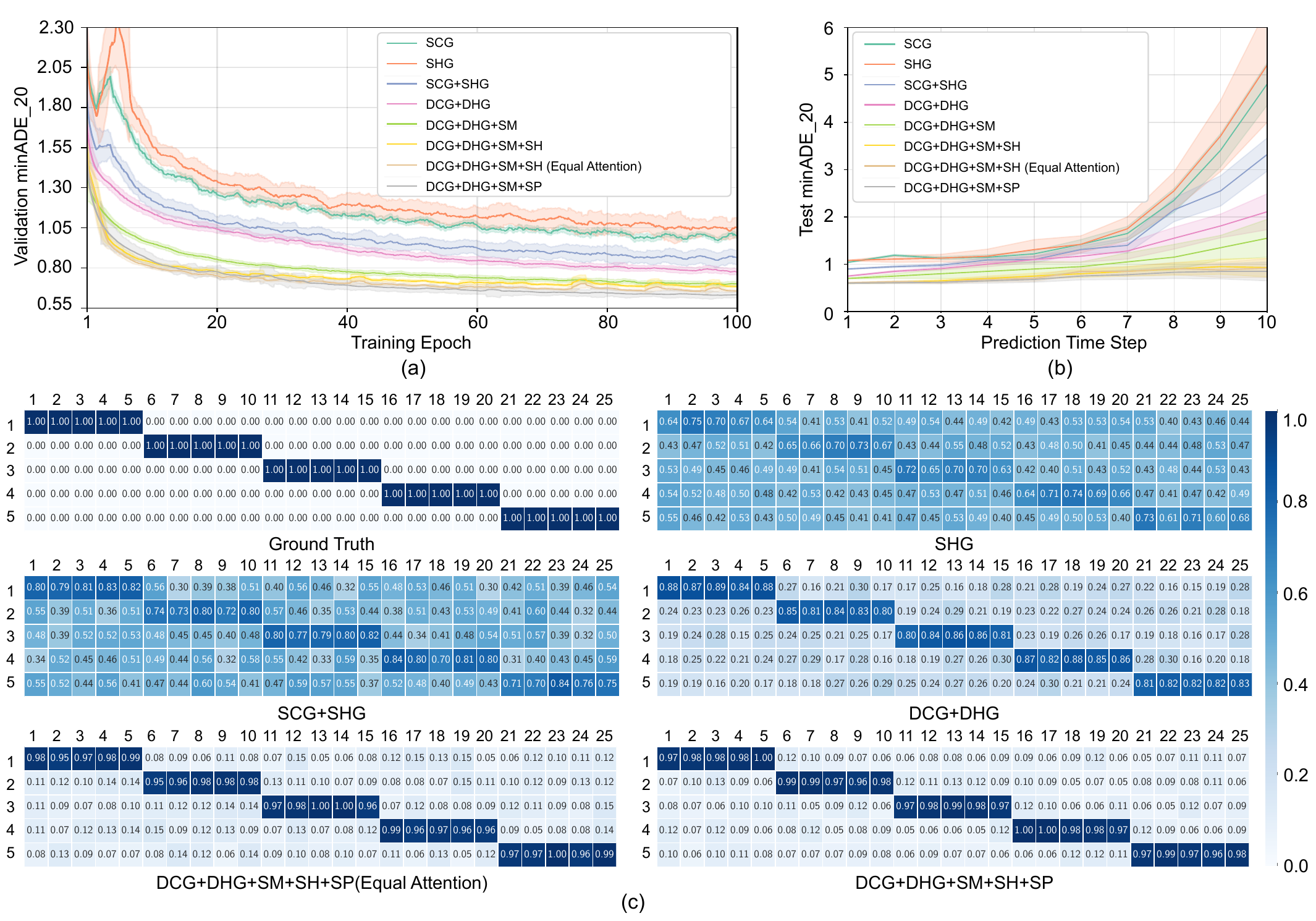}
	\vspace{-0.4cm}
	\caption{The visualizations of results on the crowd simulation dataset. (a) minADE$_{20}$ vs. training epochs on the validation dataset. The solid lines indicate the average error and the shaded areas indicate variance. (b) minADE$_{20}$ vs. prediction horizon on the testing dataset. (c) The incidence matrix of the hypergraphs inferred by different model settings. Each row represents a hyperedge and each column represents a certain agent. The numbers indicate the probability of the agent belonging to the corresponding hyperedge (i.e., group). The hyperedges of SHG and SCG+SHG shown in the figure are inferred at the first prediction step since they only infer static relations while the others are inferred at a step after the group evolution. Best viewed in color.} 
	\vspace{-0.2cm}
	\label{fig:crowd_simulation}
\end{figure*}

The RL methods under different settings were trained with distinct reward functions according to their predictive capabilities. 
Specifically, RL methods with trajectory prediction use the reward function defined in Eq. (\ref{eq:reward}) while those without trajectory prediction omit the $r_\text{pred}$ component. 
We trained the model for $2 \times 10^7$ steps with an initial learning rate of $4 \times 10^{-5}$. We used $500$ random, unseen test cases for evaluation. 
The intrusions into human personal spaces are determined based on the true future positions of pedestrians.

\subsubsection{Hardware}
All the experiments were run on an Ubuntu server equipped with an AMD Ryzen Threadripper\textsuperscript{\textregistered} 3960X 48-Core CPU and an NVIDIA RTX\textsuperscript{\textregistered} A6000 GPU.

\subsection{Dynamic Relational Reasoning for Trajectory Prediction}

\subsubsection{Quantitative Analysis}
The comprehensive comparisons of prediction results are provided in Table \ref{tab:ETH_overall}-\ref{tab:SDD_NBA}. Note that our approach achieves state-of-the-art performance across all datasets consistently while the strongest baseline methods for each dataset are different.

\paragraph{ETH-UCY and SDD datasets} We adopted the standard setting for training and testing our model to predict trajectories in the future 4.8 seconds (12 time steps) based on the historical observations of 3.2 seconds (8 time steps). 
Our approach reduces minADE$_{20}$ / minFDE$_{20}$ by 14.3\% / 16.7\% and 20.0\% / 22.7\%  on the ETH-UCY (overall) and ETH-UCY (difficult) datasets compared to the strongest baseline, respectively. Particularly, our method leads to a larger improvement for the difficult subset, indicating its advantage in modeling complex social interactions in dense scenarios. 
Our approach also achieves state-of-the-art performance on the SDD dataset. 

\paragraph{NBA dataset} We predicted the trajectories in the future 4.0 seconds (10 time steps) based on the historical observations of 2.0 seconds (5 time steps). Our approach reduces minADE$_{20}$ / minFDE$_{20}$ by 20.5\% / 25.1\% on the NBA dataset compared to the strongest baseline.

\subsubsection{Ablative Analysis}
We conducted a comprehensive ablation study on the synthetic and benchmark datasets to demonstrate the effectiveness of model components in our approach from multiple aspects.
We presented quantitative results of ablation settings along with our complete model in Fig. \ref{fig:crowd_simulation}(a)(b) and Table \ref{tab:ETH_overall}--\ref{tab:SDD_NBA}. The ablation settings are:

\begin{itemize}
    \item SCG: This variant only enables \textit{pairwise} static relational reasoning and infers a static interaction graph.
    \item SHG: This variant only enables \textit{group-wise} static relational reasoning and infers a static interaction hypergraph.
    \item SCG+SHG: This variant enables both \textit{pairwise} and \textit{group-wise} static relational reasoning.
    \item DCG+DHG: This variant enables both pairwise and group-wise \textit{dynamic} relational reasoning.
    \item DCG+DHG+SM: This variant applies a \textit{smoothness} regularization loss to the DCG+DHG model.
    \item DCG+DHG+SM+SH: This variant applies a \textit{sharpness} regularization loss to the DCG+DHG+SM model.
    \item DCG+DHG+SM+SH+SP (Equal Attention): This variant is our method without the hypergraph attention.
    \item DCG+DHG+SM+SH+SP: This is our complete method, which additionally applies a \textit{sparsity} regularization loss to the DCG+DHG+SM+SH model.
\end{itemize}

\paragraph{Group-wise relational reasoning} We show the effectiveness of group-wise relational reasoning by comparing the SCG, SHG, and SCG+SHG models.
In all the experiments, SHG performs worse than SCG, which implies that group-wise relational reasoning through hypergraphs alone cannot sufficiently model the interactions between agents. This is because pairwise interactions still dominate the multi-agent dynamics of the agents in either the same or different groups, which cannot be entirely ignored. Meanwhile, effective pairwise reasoning may help identify group behavior patterns.
Among these model variants, SCG+SHG achieves the best performance, which implies the effectiveness of integrating pairwise and group-wise relational reasoning.
SCG+SHG reduces minADE$_{20}$ by 4.2\%, 6.7\%, 5.8\%, and 7.8\% on the ETH-UCY (overall), ETH-UCY (difficult), SDD, and NBA datasets, respectively.
As shown in Fig. \ref{fig:crowd_simulation}(a) and \ref{fig:crowd_simulation}(b), SCG+SHG also leads to a significant improvement in crowd simulation and the gap becomes larger as the prediction horizon increases.

\paragraph{Dynamic Relational Reasoning} We show the effectiveness of dynamic relational reasoning by comparing the SCG+SHG and DCG+DHG models.
DCG+DHG performs better in all the experiments. 
In the crowd simulation where the group evolves significantly, the dynamic relational reasoning mechanism enhances model performance.
Since SCG+SHG cannot predict the evolution of hyperedges, the model must infer an overall group topology for the whole prediction, which limits the flexibility of reasoning.
DCG+DHG reduces minADE$_{20}$ by 8.7\%, 10.7\%, 20.8\%, and 12.2\% on ETH-UCY (overall), ETH-UCY (difficult), SDD, and NBA datasets.

\begin{table*}[!tbp]
	\setlength{\tabcolsep}{1mm}
	\caption{minADE$_{20}$ (meters) vs. Maximum number of hyperedges on the NBA dataset.}
	\vspace{-0.2cm}
	\fontsize{6}{6}\selectfont
	\centering
	\resizebox{0.8\textwidth}{!}{
		\begin{tabular}{m{1.3cm}<{\centering}|m{1.3cm}<{\centering} m{1.3cm}<{\centering} m{1.3cm}<{\centering} m{1.3cm}<{\centering} m{1.3cm}<{\centering} m{1.3cm}<{\centering} m{1.3cm}<{\centering}}
			\toprule
			\midrule
			Number & 2 & 5 & 11 & 20 & 30 & 40 & 110\\
			\midrule 
			 minADE$_{20}$ & 1.22 $\pm$ 0.01 & \textbf{1.16} $\pm$ 0.02 & 1.28 $\pm$ 0.04 & 1.29 $\pm$ 0.03 & 1.31 $\pm$ 0.04 & 1.32 $\pm$ 0.05 & 1.34 $\pm$ 0.07 \\ 
			\bottomrule
		\end{tabular}
	}
	\label{tab:NBA_hyperedge}
\end{table*}

\begin{table*}[!tbp]
	\setlength{\tabcolsep}{1mm}
	\caption{minADE$_{20}$ (meters) vs. Number of edge and hyperedge types on the NBA dataset.}
        \vspace{-0.2cm}
	\fontsize{6}{6}\selectfont
	\centering
	\resizebox{0.8\textwidth}{!}{
		\begin{tabular}{m{1.3cm}<{\centering}|m{1.3cm}<{\centering} m{1.3cm}<{\centering} m{1.3cm}<{\centering} m{1.3cm}<{\centering} m{1.3cm}<{\centering} m{1.3cm}<{\centering} m{1.3cm}<{\centering}}
			\toprule
			\midrule
			Number & 1 & 2 & 3 & 4 & 5 & 6 & 7\\
			\midrule 
			minADE$_{20}$ & 1.27 $\pm$ 0.01  & 1.23 $\pm$ 0.02 & \textbf{1.16} $\pm$ 0.02 & 1.19 $\pm$ 0.02 & 1.23 $\pm$ 0.04 & 1.27 $\pm$ 0.05  & 1.29 $\pm$ 0.04 \\
			\bottomrule
		\end{tabular}
	}
	\label{tab:NBA_hyperedge_types}
\end{table*}

\paragraph{Smoothness regularization} We show the effectiveness of smoothness regularization by comparing the DCG+DHG and DCG+DHG+SM models.
Fig. \ref{fig:crowd_simulation}(a) and \ref{fig:crowd_simulation}(b) show that DCG+DHG+SM achieves consistently better and more stable performance.
The reason is that the smoothness regularization encourages smoother evolution of graphs and hypergraphs over time and restrains abrupt changes, which leads to a more stable training process and faster convergence. It also encourages the model to learn the temporal dependency between consecutively evolving relations.
Applying the smoothness regularization loss during training reduces minADE$_{20}$ by 4.8\%, 8.0\%, 7.4\%, and 6.6\% on the ETH-UCY (overall), ETH-UCY (difficult), SDD, and NBA datasets, respectively.

\paragraph{Sharpness regularization} We show the effectiveness of the sharpness regularization by comparing DCG+DHG+SM and DCG+DHG+SM+SH models.
Fig. \ref{fig:crowd_simulation}(a) illustrates that the sharpness regularization further enhances the prediction performance in crowd simulation.
Fig. \ref{fig:crowd_simulation}(b) shows that DCG+DHG+SM+SH has a smaller variance in minADE$_{20}$ indicating that sharpness regularization also improves the model robustness to random initializations.
Applying the sharpness regularization during training reduces minADE$_{20}$ by 5.0\%, 4.3\%, 4.5\%, and 3.9\% on the ETH-UCY (overall), ETH-UCY (difficult), SDD, and NBA datasets, respectively.

\paragraph{Sparsity regularization} DCG+DHG+SM+SH+SP has the smallest prediction error among all the model settings consistently.
DCG+DHG+SM+SH+SP tends to infer sparser hypergraphs which learn more distinguishable group memberships for each agent.
The model is encouraged to learn to recognize sparse pairwise relations and clear group assignments, which creates a strong inductive bias in the message passing process through the sparse relational structures.
Applying sparsity regularization during training reduces minADE$_{20}$ by 5.3\%, 9.1\%, 3.6\%, and 5.7\% on the ETH-UCY (overall), ETH-UCY (difficult), SDD, and NBA datasets, respectively.

\begin{figure*}[!tbp]
	\centering
	\includegraphics[width=\textwidth]{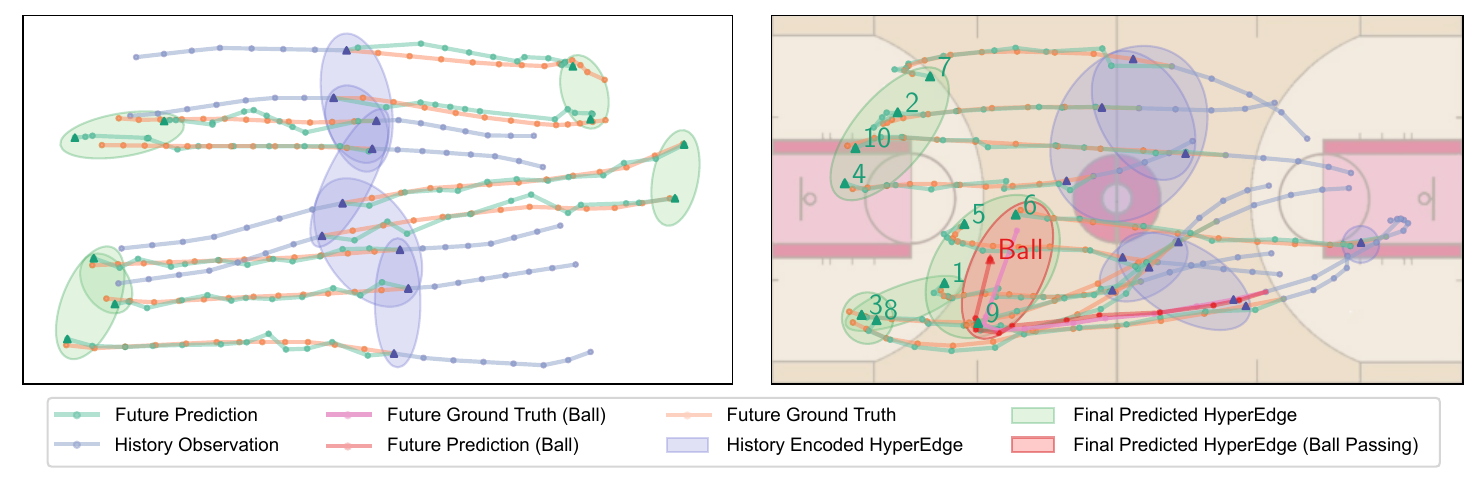}
	\caption{The visualization of trajectory prediction and relational reasoning results on the ETH-UCY dataset (left) and NBA dataset (right). The purple ellipses represent the hyperedges inferred based on historical observations while the green ones represent the hyperedges inferred at the final prediction step after evolution. Our approach captures reasonable evolving patterns of human groups in both contexts. Best viewed in color.}
	\label{fig:trajectory_plots}
\end{figure*}

\paragraph{Hyperedge numbers} In Table \ref{tab:NBA_hyperedge}, we compare the model performance with different hyperedge (i.e., group) numbers on the NBA dataset. It shows that setting 5 as the maximum number of hyperedges achieves the best performance based on a hyperparameter search. 
The model may not capture sufficient group-wise relations with only 2 hyperedges. At the same time, too many hyperedges (e.g., 11 and 110) may lead to overly dense group-wise relations that cannot capture clear group behaviors. 
In Table \ref{tab:ETH_overall}-\ref{tab:SDD_NBA}, the reported numbers of SHG show the best performance after doing a hyperparameter search. 
Based on the experimental results, we find that using a very large number of hyperedges in SHG leads to serious overfitting. Moreover, the standard deviation of the prediction error increases as the maximum hyperedge number increases because a larger number of hyperedges may lead to more local optima, which enlarges the randomness of the model performance.
These observations imply that although pairwise relations are a special type of group-wise relations where only two agents are involved, it is not sufficient to just use a large number of hyperedges to cover all the relations, which motivates our design of the integrated pairwise and group-wise relational reasoning.

\paragraph{Edge and hyperedge types} In Table \ref{tab:NBA_hyperedge_types}, we compare the model performance with different numbers of edge and hyperedge types (i.e., relation types) on the NBA dataset. 
It shows that setting 3 as the edge and hyperedge type number achieves the best performance based on a hyperparameter search.
The model may not capture a sufficient variety of pairwise or group-wise relations with only 2 edge and hyperedge types while too many types (e.g., 6) may lead to overfitting. 
We can see in Table \ref{tab:NBA_hyperedge_types} that the standard deviation of the prediction error increases as the type number increases because a larger variety of relations may lead to more local optima, which enlarges the randomness of the model performance.

\paragraph{Attention mechanisms} To validate the effectiveness of the attention mechanisms in Eq. (\ref{eq:graph_attention}) and (\ref{eq:hypergraph_attention}), we replace the attention weights in DCG+DHG+SM+SH+SP with equal weights, which shows that adopting the attention mechanism reduces minADE$_{20}$ by 14.3\%, 16.7\%, 13.8\%, and 11.5\% on the ETH-UCY (overall), ETH-UCY (difficult), SDD, and NBA datasets, respectively. This implies the benefits of modeling different contributions of agents to group-wise relations and the influence of different groups on a certain agent.

\subsubsection{Qualitative Analysis}
We visualize the predicted trajectories and the group memberships inferred in the form of hyperedges for the crowd simulation, ETH-UCY, and NBA datasets in Fig. \ref{fig:crowd_simulation}(c) and Fig. \ref{fig:trajectory_plots}, respectively.
For the crowd simulation, our approach infers accurate hyperedges with high confidence. 
First, SCG+SHG infers more clear group relations than SHG, which implies that explicitly learning pairwise relations enhances group identification.
Second, SCG+SHG infers vague hyperedges while DCG+DHG can well capture the evolved group relations, indicating the benefits of dynamic reasoning.
Third, with the proposed regularization mechanisms for relational learning, the model can infer group relations that are almost identical to the ground truth.

Fig. \ref{fig:trajectory_plots} shows the predicted trajectories and hyperedges on the ETH-UCY and NBA datasets, where our approach generates trajectory hypotheses very similar to the ground truth in both scenarios based on the inferred dynamic relations. 
In particular, our model not only captures group relations where agents within the group share similar motion patterns but also identifies group relations between a set of agents who have potential conflicts or behave adversarially.
For example, in the crowd simulation (left), our model identifies human groups (purple ellipses) who are walking in similar directions. Meanwhile, it captures the groups that are close and walking in opposite directions. These humans are inferred as another type of group, which involves interactions for collision avoidance. After the conflict period, the model infers a new set of reasonable group relations (green ellipses), which demonstrates the efficacy of dynamic reasoning.
In the NBA scenario (right), the predicted hyperedge at the final step which contains agents 6, 9, and the basketball shows that our model captures the relation of ball passing between teammates. 

\begin{table*}[!tbp]
    \setlength{\tabcolsep}{1mm}
    \caption{Social robot navigation results}
    \fontsize{6}{6}\selectfont
    \resizebox{\textwidth}{!}{
       \begin{tabular}{
            m{1.9cm}<{\raggedright} | m{1.3cm}<{\centering} m{1.3cm}<{\centering} m{1.3cm}<{\centering} m{1.0cm}<{\centering} m{1.0cm}<{\centering} m{1.3cm}<{\centering} m{1.0cm}<{\centering} m{1.3cm}<{\centering}
        }
            \toprule
            \midrule
            Method                            & \textbf{SR}$\uparrow$ & \textbf{CR}$\downarrow$ & \textbf{TR}$\downarrow$ & \textbf{NT}$\downarrow$ & \textbf{PL}$\downarrow$ & \textbf{ITR}$\downarrow$ & \textbf{SD}$\uparrow$ & \textbf{GI}$\downarrow$\\
            \midrule 
            ORCA \cite{van2011reciprocal}      & 80.77$\pm$1.93$\%$ & 18.59$\pm$1.82$\%$ & 0.64$\pm$0.10$\%$ & 14.98$\pm$0.09 & 20.71$\pm$0.36 & 18.28$\pm$0.82$\%$ & 0.36$\pm$0.02 & 22.45$\pm$1.51$\%$  \\
            DSOA\cite{huber2022avoiding}      & 90.25$\pm$0.98$\%$ &  4.20$\pm$0.68$\%$ & 5.55$\pm$0.32$\%$ & 16.02$\pm$0.11 & 23.15$\pm$0.42 & 27.41$\pm$1.01$\%$ & 0.18$\pm$0.02 & 38.41$\pm$2.12$\%$  \\
            RL (no prediction)                 & 83.02$\pm$2.41$\%$ & 16.63$\pm$2.33$\%$ & 0.35$\pm$0.08$\%$ & 16.39$\pm$0.16 & 22.56$\pm$0.21 &  7.21$\pm$0.27$\%$ & 0.32$\pm$0.03 &  9.91$\pm$0.20$\%$  \\
            RL (constant velocity)             & 84.11$\pm$2.55$\%$ & 15.63$\pm$1.31$\%$ & 0.26$\pm$0.05$\%$ & 15.86$\pm$0.18 & 21.71$\pm$0.20 &  4.01$\pm$0.11$\%$ & 0.29$\pm$0.02 &  8.98$\pm$0.12$\%$  \\
            Group-Aware \cite{katyal2022learning} & 91.68$\pm$1.54$\%$ &  7.93$\pm$1.45$\%$ & 0.39$\pm$0.06$\%$ & 15.27$\pm$0.11 & 21.08$\pm$0.19 &  3.42$\pm$0.23$\%$ & 0.33$\pm$0.03 &  8.79$\pm$0.21$\%$ \\
            CrowdNav++ \cite{liu2023intention} & 93.21$\pm$1.31$\%$ &  6.66$\pm$1.30$\%$ & 0.13$\pm$0.03$\%$ & 14.81$\pm$0.07 & 20.37$\pm$0.15 &  3.11$\pm$0.10$\%$ & 0.38$\pm$0.02 &  8.66$\pm$0.14$\%$  \\
            \midrule                                       
            Ours (w/o DHG+GR)                  & 94.57$\pm$1.21$\%$ &  5.33$\pm$1.33$\%$ & 0.10$\pm$0.03$\%$ & 14.88$\pm$0.08 & 20.39$\pm$0.14 & 3.12$\pm$0.10$\%$ & 0.36$\pm$0.03 & 8.51$\pm$0.18$\%$  \\
            Ours (w/o GR)                      & 96.08$\pm$0.61$\%$ &  3.81$\pm$0.55$\%$ & 0.11$\pm$0.03$\%$ & \textbf{14.78$\pm$0.10} & \textbf{20.33$\pm$0.15} & 3.09$\pm$0.10$\%$ & 0.37$\pm$0.02 & 8.43$\pm$0.12$\%$  \\
            Ours                               & \textbf{96.31$\pm$0.53$\%$} & \textbf{3.59$\pm$0.50$\%$} & \textbf{0.10$\pm$0.03$\%$} & 14.92$\pm$0.15 & 20.53$\pm$0.13  & \textbf{ 3.08$\pm$0.08$\%$} & \textbf{0.40$\pm$0.02} & \textbf{ 5.21$\pm$0.11$\%$} \\
            \midrule
            Oracle                             & 97.28$\pm$0.46$\%$ & 3.07$\pm$0.41$\%$ & 0.05$\pm$0.05$\%$ & 14.57$\pm$0.24 & 20.05$\pm$0.13 & 3.01$\pm$0.09$\%$ & 0.42$\pm$0.03 &  4.29$\pm$0.04$\%$  \\
            \bottomrule
        \end{tabular}
    }
    \label{tab:rl_result}
\end{table*}

\subsection{Social Robot Navigation}

\subsubsection{Quantitative Analysis}

\begin{figure*}[!tbp]
	\centering
	\includegraphics[width=0.95\textwidth]{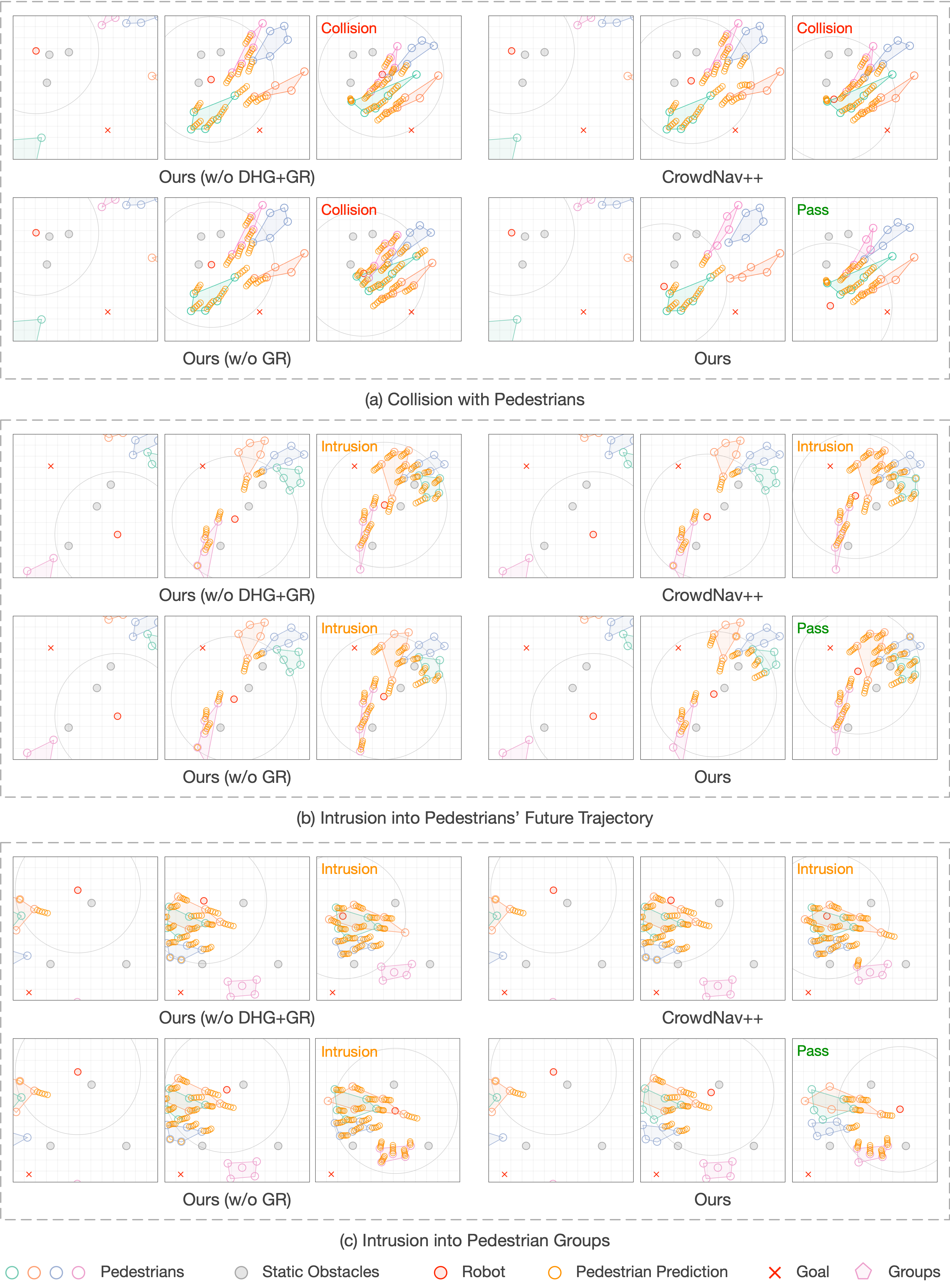}
        \vspace{-0.2cm}
	\caption{The visualization of social robot navigation results of different methods in three representative test cases: (a) collision with pedestrians; (b) intrusion into pedestrians' future trajectories; and (c) instruction into pedestrian groups. In each case, we select three representative frames to show the evolution of the dynamic scenes. Best viewed in color.}
	\label{fig:robot_navigation_plots}
\end{figure*}

To demonstrate the effectiveness of our social robot navigation method with dynamic relational reasoning, we provide a comprehensive comparison between our method and baselines in terms of a series of standard evaluation metrics.
We also include an Oracle method that employs the true future trajectories of humans for RL policy learning, which serves as a performance upper bound for our method. Note that this Oracle method is not practically deployable and only used for the comparative study.
The experimental results are shown in Table \ref{tab:rl_result}. 

By employing our trajectory prediction model with relational reasoning, the social navigation method significantly outperforms the rule-based method ORCA \cite{van2011reciprocal},
the dynamical system-based obstacle avoidance (DSOA) algorithm \cite{huber2022avoiding},
the non-predictive method RL (no prediction), the predictive baseline RL (constant velocity),
Group-Aware \cite{katyal2022learning},
as well as CrowdNav++ \cite{liu2023intention} across all the evaluation metrics. 
The improvement is evident in achieving a higher success rate (SR), lower collision rate (CR) and timeout rate (TR), and shorter navigation time (NT) and path lengths (PL). 
Meanwhile, our method notably enlarges the social distance from humans and reduces improper intrusions into human personal or group spaces, as indicated by SD, ITR, and GI. 
This enables the robot to find safer navigation trajectories and minimize interference with pedestrians.
Although the DSOA algorithm \cite{huber2022avoiding} effectively prevents collisions between robots and individual humans or small numbers of pedestrians, its disregard for group dynamics may cause the robot to inadvertently enter the center of a group. This positioning can lead to longer navigation paths as the robot must then navigate out of the group, potentially resulting in unavoidable collisions and unsafe navigation conditions. Moreover, the Group-Aware model \cite{katyal2022learning}, while considering group formation of pedestrians, exhibits limited performance when interacting with multiple surrounding dynamically evolving groups, which may lead to suboptimal actions.
Our method achieves comparable performance to the Oracle method which has access to the ground truth human trajectories, showing the effectiveness of our prediction model.

\subsubsection{Ablative Analysis}
Here we show the benefits brought by group-wise relational reasoning and group-based reward.

\paragraph{Group-wise reasoning} 
Ours (w/o DHG+GR) in Table \ref{tab:rl_result} represents a variant of our method with only pairwise reasoning and hence it disables group-based reward.
Due to the lack of understanding of group relations, this variant leads to a higher CR and lower SD because the prediction model may not predict group behaviors accurately, which degrades the safety of robot navigation. Meanwhile, it causes a higher ITR and GI during navigation, which may make the surrounding humans less comfortable.
Although it achieves a slightly shorter NT and PL (i.e., higher efficiency), the improvement is brought by the cost of social appropriateness, which is not desired.

\paragraph{Group reward} 
Ours (w/o GR) in Table \ref{tab:rl_result} represents a variant of our method without applying the group-based reward for RL policy learning. It still uses group-wise reasoning for trajectory prediction.
The results show that without explicitly considering group relations in the reward function design, the robot policy still leads to a higher CR, ITR, GI, and a lower SD, although it achieves a comparable SR to our complete method.
The NT and PL of our method are slightly larger because the policy tends to choose a safer and socially compliant trajectory which may take detours during navigation, which is acceptable and even desired in practice.

\subsubsection{Qualitative Analysis}

We visualize several test cases and compare the robot trajectories planned by different variants of our methods and the state-of-the-art baseline (i.e., CrowdNav++ \cite{liu2023intention}) in Fig. \ref{fig:robot_navigation_plots}.
We mainly highlight the impact of group-wise relational reasoning and group-based reward on RL-based social robot navigation.

In these cases, Ours (w/o GR) predicts pedestrians' future trajectories more accurately than Ours (w/o DHG+GR) and CrowdNav++ consistently, which implies the effectiveness of group-wise reasoning for interactive behavior prediction. 
However, in Fig. \ref{fig:robot_navigation_plots}(a), all the baseline methods cause a collision with pedestrians because they navigate the robot into a very dense area. In contrast, Ours plans a safer and more socially compliant trajectory thanks to the integration of group-based rewards for policy learning, which avoids getting into dense areas while maintaining good efficiency.
Similarly, in Fig. \ref{fig:robot_navigation_plots}(b), all the baseline methods lead to an intrusion into future trajectories of pedestrians, which may interfere with human intentions and violate social norms.
In Fig. \ref{fig:robot_navigation_plots}(c), all the baseline methods cause an intrusion into human groups while Ours accurately captures the groups and navigates the robot through the interval between different groups safely.
In all cases, our method considers group dynamics, which not only prevents collisions with individual pedestrians but also reduces intrusions into the human group as a whole. 

\begin{figure}[!tbp]
	\centering
	\includegraphics[width=\columnwidth]{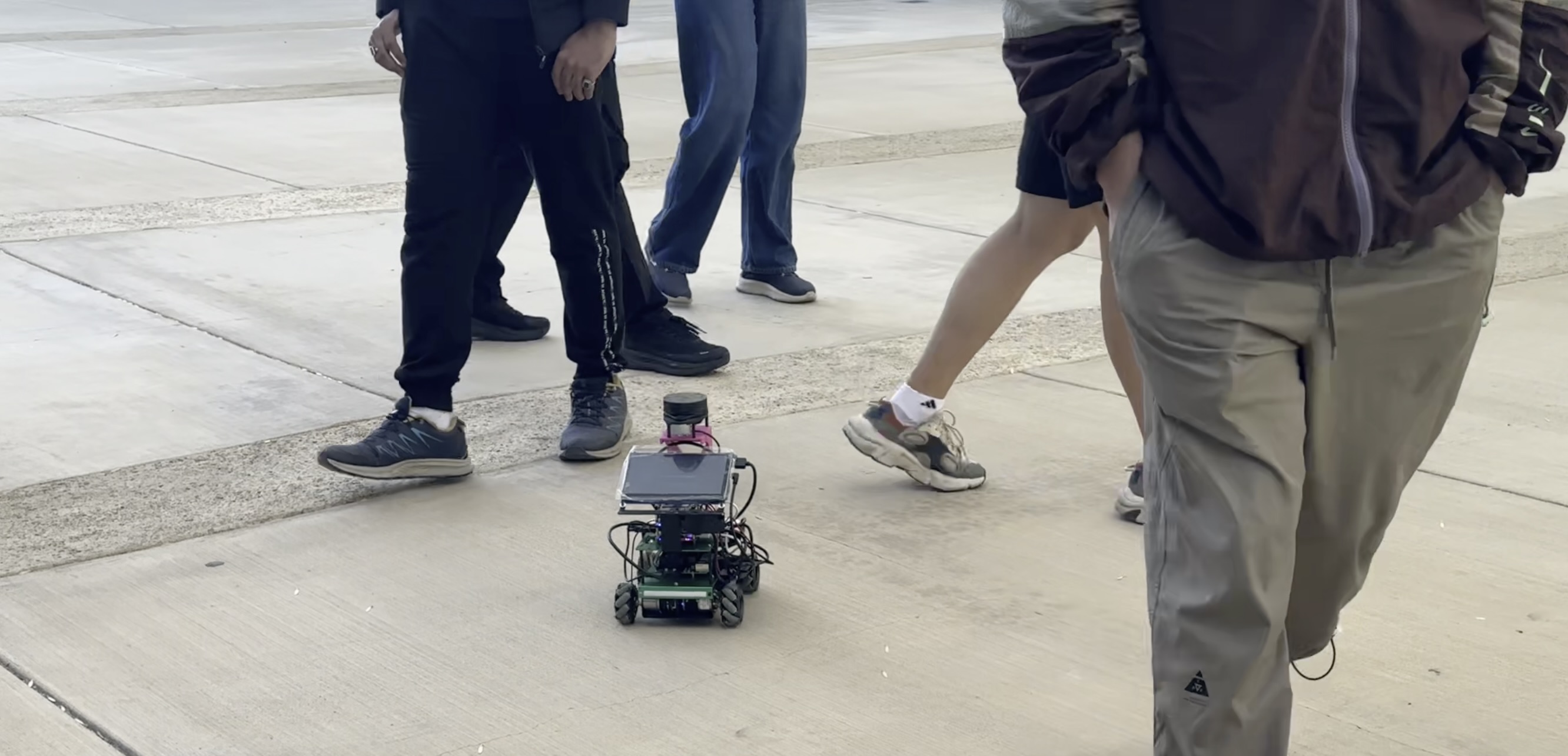}
	\caption{A screenshot of our experiment video showing that the robot equipped with our method can interact properly with dense crowds with natural moving patterns in real-world scenarios and generate safe paths through crowds.}
	\label{fig:real_robot}
\end{figure}

\subsubsection{Real-World Experiments}
We deploy our model, trained in the simulation environments, directly on a RosMaster X3 robot equipped with four Mecanum wheels, which enables independent speed control in both horizontal and lateral directions. Different from ideal holonomic kinematics simulations, the RosMaster X3 robot cannot handle sudden changes in velocity. 
Therefore, we implement smoothness constraints to limit the maximum change in velocity, setting the maximum acceleration to \SI{1}{m/s^{2}} and the maximum velocity to \SI{0.8}{m/s}. 
We apply direct clipping after receiving the velocities generated by our RL agent.
We use a 2D LiDAR, RPLIDAR-A1, to detect human positions through a learning-based human detector, DR-SPAAM \cite{jia2020dr}.
The robot navigates on a large outdoor terrace with natural pedestrians, which covers an area larger than \SI{10}{m}$\times$\SI{12}{m}. 
In our environment, pedestrians walk at a natural speed with no constraints as shown in Fig. \ref{fig:real_robot}.

We select goal points for the robot close to human crowds to encourage maximum interactions. In our experiments, the robot is surrounded by up to 20 pedestrians in complex, dynamic scenarios.
As shown in our experiment videos, and the robot successfully navigates to its goal while making appropriate decisions without collisions. In addition, we intentionally position humans to block the robot's path, and the robot responds promptly to avoid collisions, indicating that our navigation policy adapts well to various pedestrian motion patterns and interacts with
real humans proactively despite noises from the perception
module and robot dynamics.

\section{Conclusion} \label{sec:conclusion}

In this paper, we present a systematic dynamic relational reasoning framework that infers pairwise and group-wise relations, which proves to be effective for multi-agent trajectory prediction and social robot navigation.
In addition to inferring the conventional graphs that model pairwise relations between interacting agents, we propose to infer relation hypergraphs to model group-wise relations that exist widely in real-world scenarios such as human crowds. 
Different hyperedge types represent distinct group-wise relations. 
We employ a dynamic reasoning mechanism to capture the evolution of underlying relational structures.
We propose effective mechanisms to regularize the smoothness of relation evolution, the sharpness of the learned relations, and the sparsity of the learned graphs and hypergraphs, which not only enhances training stability but also reduces prediction error.
The proposed approach infers reasonable relations and achieves state-of-the-art performance on both synthetic crowd simulations and multiple benchmark datasets.
Moreover, we present a deep reinforcement learning framework for social robot navigation that incorporates relational reasoning and trajectory prediction in a systematic way.
It achieves state-of-the-art performance in terms of safety, efficiency, and social appropriateness compared with advanced baselines.
A limitation of this work is that although our method outperforms the strongest baseline by a large margin, it still cannot achieve a zero collision rate in very dense scenarios. Therefore, in future work, we will investigate how to enable a provably safety guarantee. 

\bibliographystyle{IEEEtran}
\bibliography{ref}

\begin{IEEEbiography}[{\includegraphics[width=1in,height=1.25in,clip,keepaspectratio]{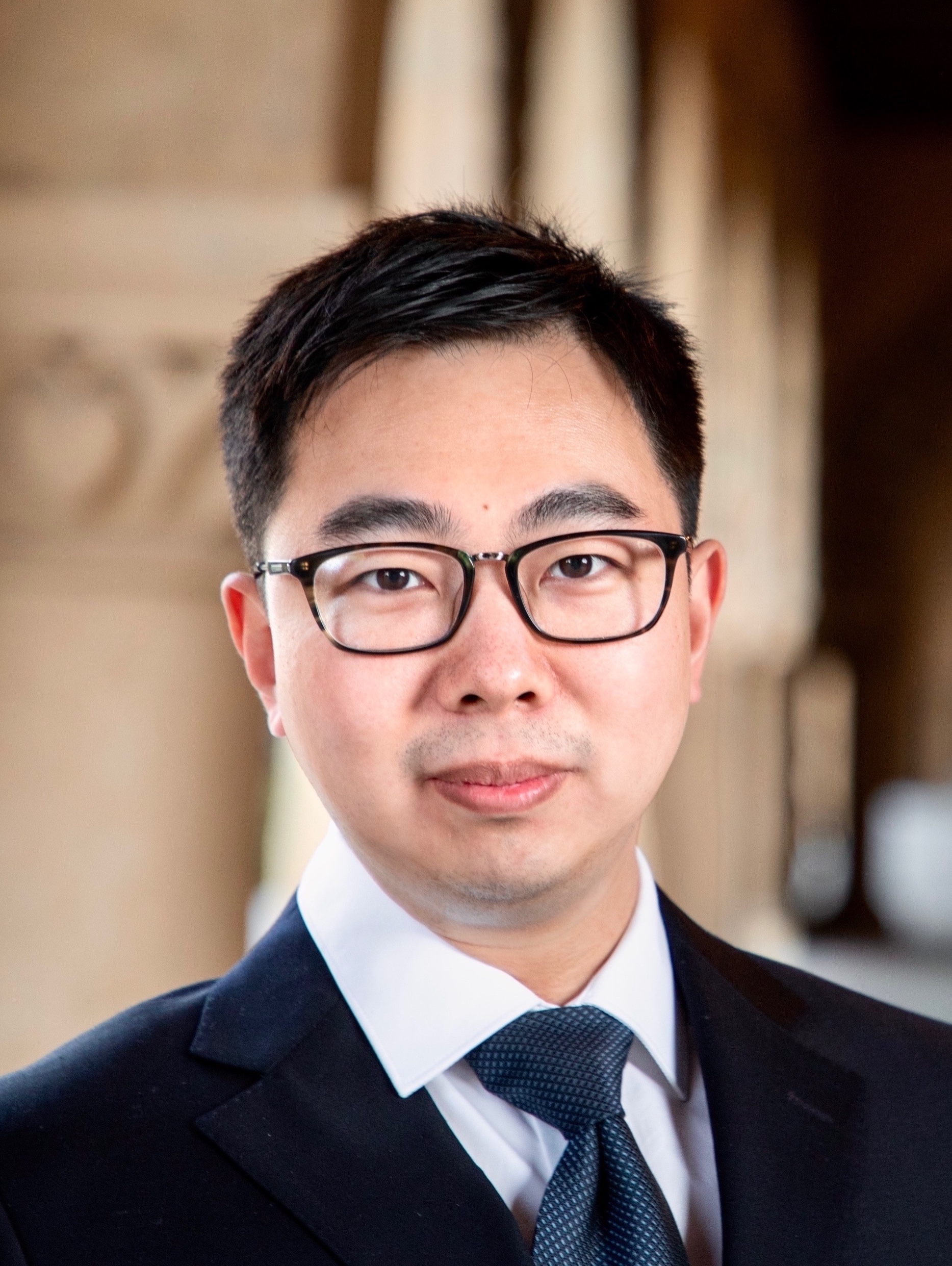}}]{Jiachen Li}
(Member, IEEE) is currently an assistant professor in the Department of Electrical and Computer Engineering and the Department of Computer Science and Engineering (by courtesy) at the University of California, Riverside, USA. 
He is the director of the Trustworthy Autonomous Systems Laboratory (TASL).
Prior to joining the faculty, he was a postdoctoral scholar at Stanford University.
He received his Ph.D. from the University of California, Berkeley in 2021. He received a B.E. degree from Harbin Institute of Technology, China in 2016. 
His research interest lies at the broad intersection of robotics, trustworthy AI, reinforcement learning, control and optimization, and their applications to intelligent autonomous systems, particularly in human-robot interactions and multi-agent systems. Dr. Li was recognized as an RSS Robotics Pioneer in 2022 and an ASME DSCD Rising Star in 2023. He serves as an associate editor or a reviewer for multiple journals and conferences. He has organized multiple workshops on robotics, machine learning, computer vision, and intelligent transportation systems. 
\end{IEEEbiography}

\begin{IEEEbiography}[{\includegraphics[width=1in,height=1.25in,clip,keepaspectratio]{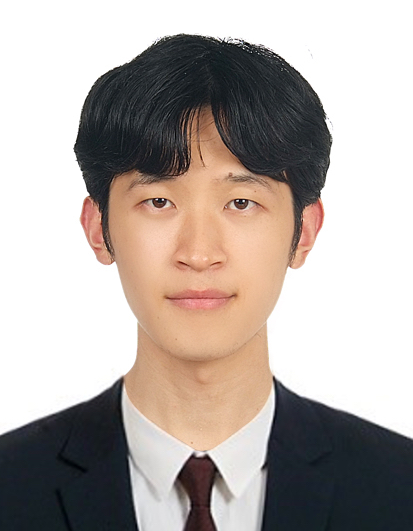}}]{Chuanbo Hua} received his B.S. degree in Mathematics from the Harbin Institute of Technology, China, in 2020. He was a Computer Science Engineering exchange researcher at Pohang University of Science and Technology (POSTECH), Republic of Korea, in 2019. He then completed his M.S. degree in Industrial and Systems Engineering at the Korea Advanced Institute of Science and Technology (KAIST), Republic of Korea, in 2022. Currently, he is a Ph.D. candidate in the System Intelligence Laboratory, part of the Department of Industrial and Systems Engineering at KAIST. His research interests are primarily focused on combinatorial optimization, reinforcement learning, and dynamic physical process simulation.
\end{IEEEbiography}

\begin{IEEEbiography}[{\includegraphics[width=1in,height=1.25in,clip,keepaspectratio]{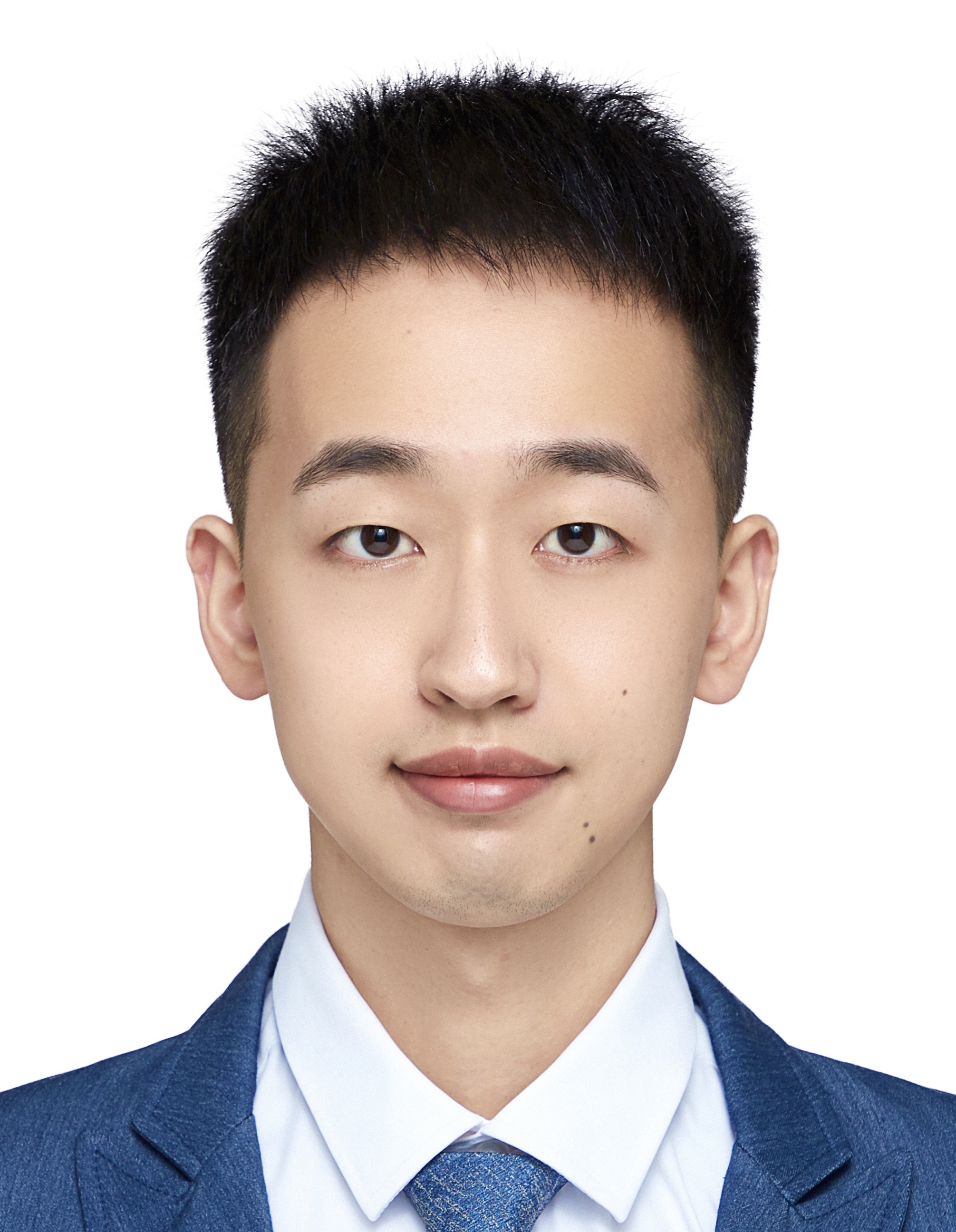}}]{Jianpeng Yao} received his B.S. and M.S. degrees in Electrical Engineering from Huazhong University of Science and Technology in 2019 and 2022, respectively. He worked as an algorithm engineer in the autonomous driving industry, where he focused on learning-based trajectory prediction and generation. Currently, he is a Ph.D. student at the University of California, Riverside. His research interests include reinforcement learning, trajectory planning, and trustworthy AI, with applications to robotics and autonomous systems.
\end{IEEEbiography}


\begin{IEEEbiography}[{\includegraphics[width=1in,height=1.25in,clip,keepaspectratio]{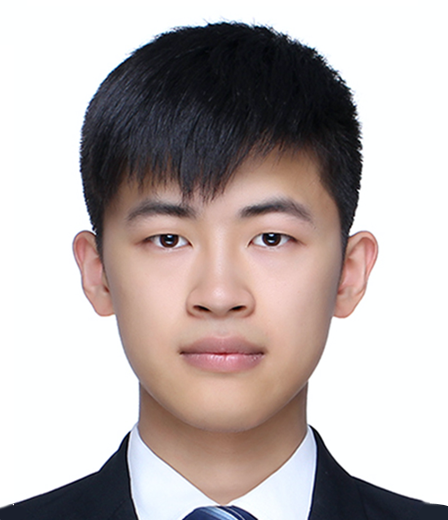}}]{Hengbo Ma} received his Ph.D. degree from the Department of Mechanical Engineering at the University of California, Berkeley in 2023. 
During his Ph.D. study, Hengbo Ma conducted research on human behavior understanding for robotics and autonomous driving. 
Before that, he received a B.E. degree from the Department of Control Science and Engineering, Harbin Institute of Technology, China in 2017.
His research interest lies in the intersection of control, decision making, optimization, machine learning and artificial intelligence with their applications in robotics and autonomous systems. 
\end{IEEEbiography}

\begin{IEEEbiography}[{\includegraphics[width=1in,height=1.25in, clip,keepaspectratio]{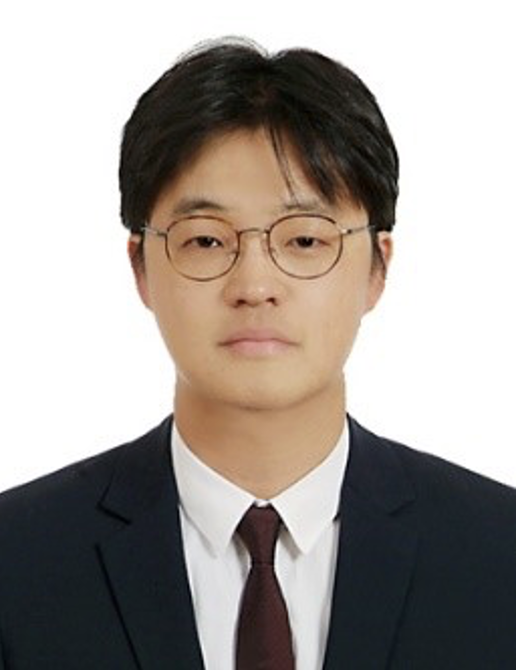}}]{Jinkyoo Park} is currently an associate professor of Industrial and Systems Engineering and adjoint professor of Graduate School of Artificial Intelligence at Korea Advanced Institute of Science and Technology (KAIST), Republic of Korea. He received his B.S. degree in Civil and Architectural Engineering from Seoul National University in 2009, an M.S. degree in Civil, Architectural and Environmental Engineering from the University of Texas Austin in 2011, an M.S. degree in Electrical Engineering from Stanford University in 2015, and a Ph.D. degree in Civil and Environmental Engineering from Stanford University in 2016. His research goal is to explore the potential of the various machine-learning approaches for improving complex decision-making methods in optimization, optimal control, and game theory.
\end{IEEEbiography}

\begin{IEEEbiography}[{\includegraphics[width=1in,height=1.25in,clip,keepaspectratio]{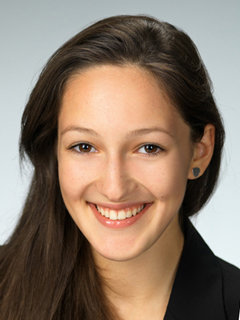}}]{Victoria M. Dax,} currently pursuing her Ph.D. at Stanford Intelligent Systems Laboratory, is focused on integrating niche advancements from other deep learning methods, such as Reinforcement Learning and Variational Autoencoders, into Graph Neural Networks. Before starting her doctoral studies, Victoria worked in the autonomous driving industry, where she developed a practical understanding of applying intelligent systems. Victoria began her academic path at ETH Zurich, earning a Bachelor of Science in Mechanical Engineering in 2016. She furthered her education at Stanford University, where she completed a Master's degree in Aeronautics and Astronautics in 2018.
\end{IEEEbiography}

\begin{IEEEbiography}[{\includegraphics[width=1in,height=1.25in,clip,keepaspectratio]{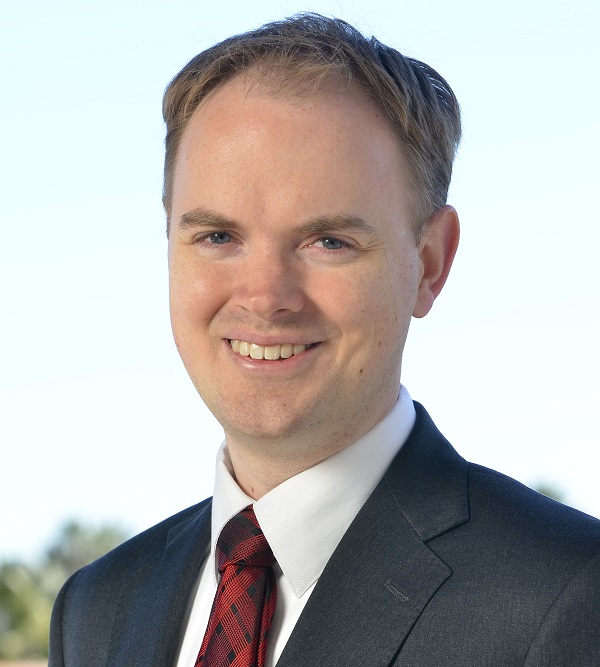}}]{Mykel J. Kochenderfer}
(Senior Member, IEEE) is an Associate Professor of Aeronautics and Astronautics at Stanford University. He is the director of the Stanford Intelligent Systems Laboratory (SISL), conducting research on advanced algorithms and analytical methods for the design of robust decision making systems. Prior to joining the faculty in 2013, he was at MIT Lincoln Laboratory where he worked on aircraft collision avoidance for manned and unmanned aircraft. He received his Ph.D. from the University of Edinburgh in 2006. He received B.S. and M.S. degrees in computer science from Stanford University in 2003. He is an author of the textbooks \textit{Decision Making under Uncertainty: Theory and Application} (MIT Press, 2015), \textit{Algorithms for Optimization} (MIT Press, 2019), and \textit{Algorithms for Decision Making} (MIT Press, 2022).

\end{IEEEbiography}
\vfill

\end{document}